\documentclass[journal]{IEEEtran}
\usepackage{amssymb}
\usepackage{amsmath}
\usepackage{pdflscape}
\usepackage{array,multirow}
\usepackage{rotating}
\usepackage{graphicx}
\usepackage{subfig}
\usepackage{tabularx}
\usepackage{multirow}
\ifCLASSINFOpdf
\else
\fi
\begin{document}
%
\title{Support Vector Regression via a Combined Reward Cum Penalty Loss Function}
%
%
%

\author{Pritam Anand, Reshma Rastogi 
        and Suresh Chandra.
\thanks{ P. Anand and R. Rastogi are with the Department of
Computer Science, South Asian University, New-Delhi-110021, India. E-mail: reshma.khemchandani@sau.ac.in, ltpritamanand@gmail.com.}
\thanks{ Suresh Chandra is the Ex-faculty of the Department of Mathematics, Indian Institute of Technology, Hauz-Khas, New-Delhi- 110016, India.  E-mail:sureshiitdelhi@gmail.com.}
\thanks{}}

%
%

\markboth{Preprint}%
{Shell \MakeLowercase{\textit{et al.}}: Bare Demo of IEEEtran.cls for IEEE Journals}
%



\maketitle

\begin{abstract}
In this paper, we introduce a novel combined reward cum penalty loss function to handle the regression problem. The proposed combined reward cum penalty loss function penalizes the data points which lie outside the $\epsilon$-tube of the regressor and also assigns reward for the data points which lie inside of the $\epsilon$-tube of the regressor. The combined reward cum penalty loss function based regression (RP-$\epsilon$-SVR) model has several interesting properties which are investigated in this paper and are also supported with the experimental results. 
\end{abstract}

%
\IEEEpeerreviewmaketitle

\section{Introduction}
Past few decades have witnessed the evolution of the Support Vector Regression (SVR) models (Vapnik et al. \cite{svr1}, Drucker et al. \cite{svr2}, Smola and Scholkopf \cite{svr3},  Gunn \cite{GUNNSVM}, Vapnik \cite{statistical_learning_theory}) as a promising tool for handling the problem of function approximation. 
It has been successfully used in a wide variety of applications, e.g.  \cite{wind}  to \cite{timeseries2}. SVR models have also been extended in non-parallel framework e.g. \cite{twsvr1} to \cite{twsvr2}.

Given a training set $T=\{(x_i,y_i): x_i \in \mathbb{R}^n, y_i \in \mathbb{R}, i=1,2,.. ,l\}$,  a typical SVR model determines a regressor $~f(x)= w^T\phi(x)+ b~$, $w \in  \mathbb{R}^n$,$b \in \mathbb{R}$ in feature space for predicting the response of a unseen test point. It uses the training set to minimize the empirical risk. In addition to this, it also minimizes a regularization term  in its optimization problem for minimizing the structural risk. 

There exist several SVR models in the literature. These models commonly use  different types of loss functions to measure their empirical risk along with different types of  regularizations. Some of SVR models which uses the convex loss function are as follows.
\begin{enumerate}
	\item[(i)] The standard $\epsilon$-SVR model (Drucker et al. \cite{svr2}) uses the $\epsilon$-insensitive loss function to measure the empirical risk with the regularization term $\frac{1}{2} w^Tw$.
	\item[(ii)] The standard Least Squares Support Vector Regression (LS-SVR) model  (Suykens and Vandewalle  \cite{LSSVR2}) uses the quadratic loss function to measure the empirical risk along  with the regularization term $\frac{1}{2} w^Tw$. 
	\item[(iii)] Maximum Likelihood Optimal and Robust Support Vector Regression model (Karal  \cite{incoshloss}) uses the lncosh loss function to measure the empirical risk with the regularization term $\frac{1}{2} w^Tw$.
	\item[(iv)] Huber loss function based SVR (Gunn \cite{GUNNSVM} ) uses the Huber loss function to measure the empirical risk along with the regularization term $\frac{1}{2} w^Tw$ .
	\item[(v)] $L_1$-norm SVR (Tanveer et al. \cite{l1normsvr}) uses the $\epsilon$-insensitive loss function for measuring the empirical risk with the $L_1$-norm regularization term $\frac{1}{2}||w||_1$ .
	\item[(vi)] Large-margin Distribution Machine based Regression (LDMR) model (Rastogi et al. \cite{LDMR}) uses a linear combination of the $\epsilon$-insensitive loss function and the quadratic loss function for measuring the empirical risk with the $L_2$-norm regularization.
	\item[(vii)] Penalizing-$\epsilon$-generalized SVR (Anand et al., \cite{pensvr}) uses the generalized $\epsilon$-loss function to measure the empirical risk along with the regularization term $\frac{1}{2}w^Tw$. 
\end{enumerate}
 Apart from the above convex loss functions, some non-convex loss functions have also been used by researchers in SVR models. Some important of them are smooth Ramp loss function ( Zhao and Sun, \cite{flexible_loss} ) non convex least square loss  function (Wang and Zhong, \cite{nonconvex_least}), non convex generalized loss function (Wang et al., \cite{nonconvex_genralized}) generalized quantile loss (Yang et al., \cite{yang2019robust}) and rescaled expectile loss (Yang et al., \cite{yang2020robust}). However, the optimization problem of non convex loss function based SVR models are algorithmically complex and computationally expensive.

\begin{figure*}
	\centering
	\includegraphics[width=0.8\linewidth, height=0.3\textheight]{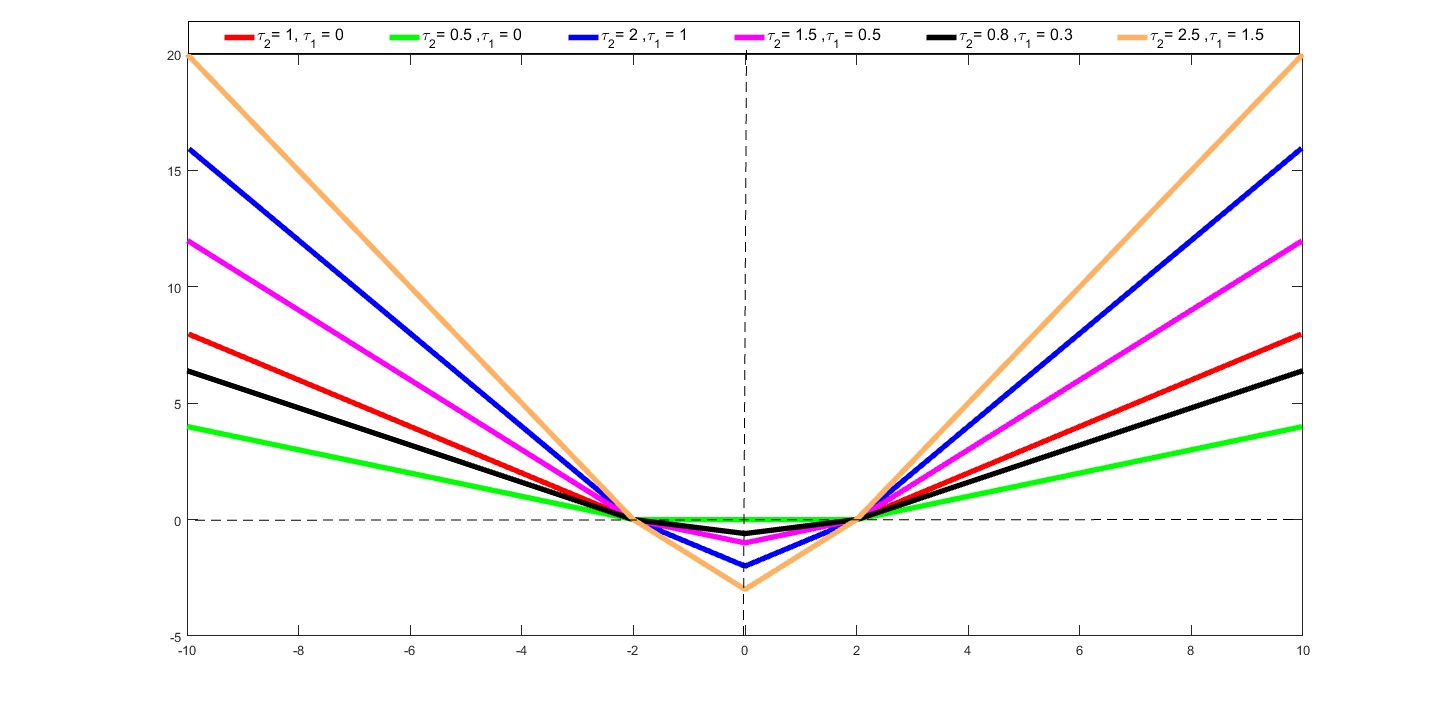}
	\caption{ Plot of the reward cum penalty loss with $\epsilon$ = 2 and different value of $\tau_2$ and $\tau_1$. The proposed loss functions reduce to popular $\epsilon$-insensitive loss function with $\tau_2=1$ and $\tau_1=0$ (red line).   }
	\label{:RewardPenalitylossfunction}
\end{figure*} 
 
A  particular choice of loss function in a SVR model enables it to obtain optimal estimate for a particular type of noise model. For example, the use of quadratic loss function in LS-SVR model enables it to perform optimal for the normal noise. But, an ideal SVR model is expected to be robust i,e. it should perform well without bothering the nature of noise present in data. Apart from this, it should perform optimal for a wide family of noise distributions and manage to obtain sparse solution as well.


In existing SVR models, the  standard $\epsilon$-SVR model is the most popular one. It is because of the fact that, it is a robust SVR model and can also manage to obtain the sparse solution.

 The standard $\epsilon$-SVR model minimizes the regularization $\frac{1}{2} w^Tw$ to make the estimated regressor as flat as possible along with $\epsilon$-insensitive loss function to minimize the empirical risk.  The $\epsilon$-insensitive loss function is given as follows
\begin{eqnarray}
L _{\epsilon}(y_i,x_i,f(x_i)) = \begin{cases}
|y_i-f(x_i)|-\epsilon,~if~|y_i-f(x_i)|\geq\epsilon,\\
0 ~~~otherwise,
\end{cases}
\end{eqnarray}  where $\epsilon \geq 0$ is a parameter. 
The use of $\epsilon$-insensitive loss function in standard $\epsilon$-SVR model makes it to ignore those data points which lie inside the $\epsilon$-tube of the regressor $f(x)$. The data points which lie outside the $\epsilon$-tube are penalized in the optimization problem to bring them close to the $\epsilon-$tube. These data points along with the data points lying on the boundary of the $\epsilon$-tube of $f(x)$ constitute `support vectors' which only decide the orientation and position of the regressor $f(x)$. 

The use of the $\epsilon$-insensitive loss function in the $\epsilon$-SVR model enables it to  obtain a robust and sparse solution.  But, it also causes it to lose  most of the information contained in the training set in the sense that data points lying inside of the $\epsilon$-tube are ignored in the construction of regressor. Further, the performance of the $\epsilon$-SVR model is subjected to having a right choice of the value of the $\epsilon$. A wrong choice of $\epsilon$ may result in the loss of significant part of the information contained in the training set and can lead to poor generalization ability.

\begin{figure}
	\centering
	\includegraphics[width=1.0\linewidth, height=0.2\textheight]{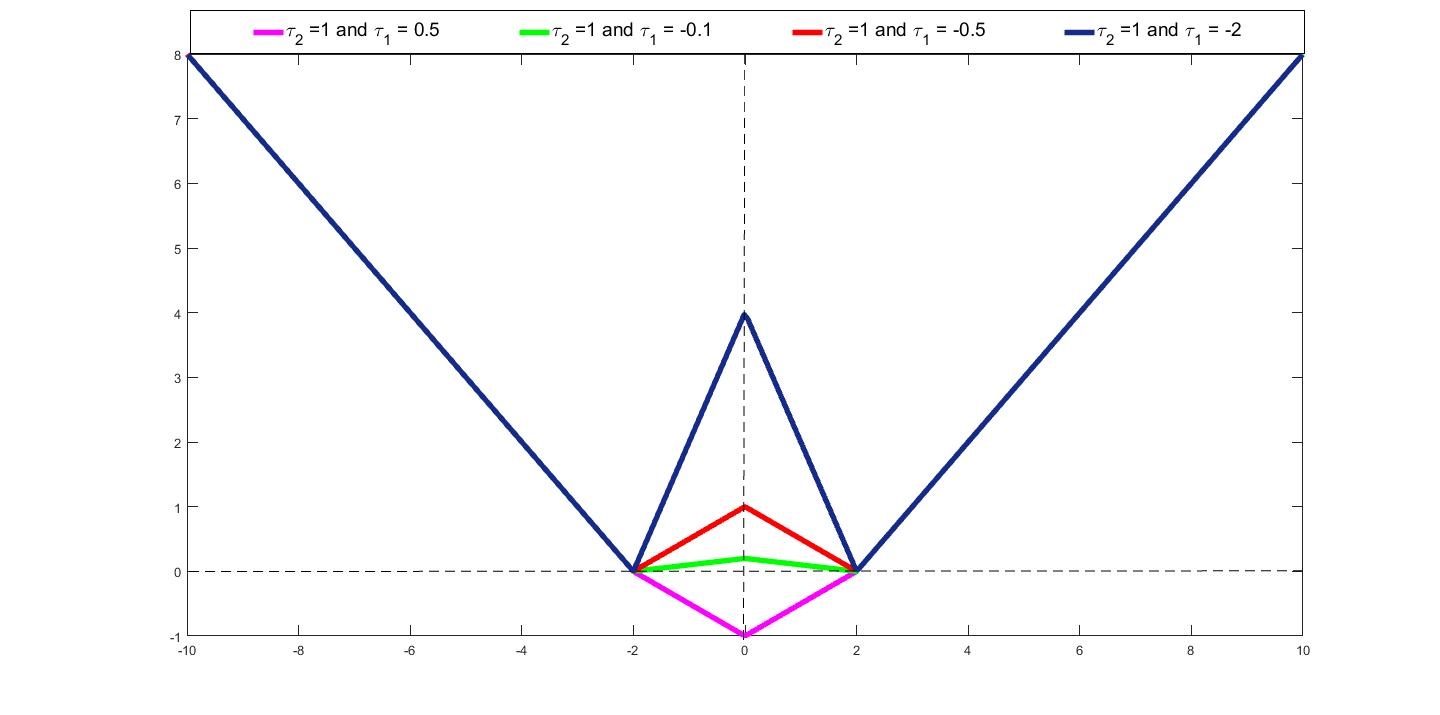}
	\caption{ Plot of the reward cum penalty loss function with $\epsilon$ = 2, $\tau_2 = 1$ and different value of $\tau_1$. The proposed reward-loss functions loses its convexity with $\tau_1 < 0$.}
	\label{:RewardPenalitylossfunctionwithnegativevalue}
\end{figure}

 We require a SVR model which can properly use the training set and can also preserve the elegance of the $\epsilon$-SVR model simultaneously. Taking motivation from this, we propose a new convex loss function termed as `reward cum penalty loss function'. Unlike the existing loss function, the proposed reward cum penalty loss function can take both positive and negative values. Here, a positive value represents `penalty' and a negative value represent `reward'.  It penalizes those data points which do not lie on the desired location and rewards those data points which lie on the desired location. The proposed reward cum penalty loss function is given by    
\begin{eqnarray}
RP_{\tau_1, \tau_2 ,\epsilon}(u) =  \max(~\tau_2(|u|-\epsilon), ~\tau_1(|u|-\epsilon)~),
\end{eqnarray}
where $\tau_2$, $\tau_1$ and $\epsilon \geq 0$ are parameters.    
For the regression training set $T= \{(x_i,y_i): x_i \in \mathbb{R}^n, y_i \in \mathbb{R},~ i=1,2,..,l\}$, the above proposed loss function can be used to measure the empirical error as follow
{\small  \begin{eqnarray}
	&  \hspace{-50mm}RP_{\tau_1, \tau_2 ,\epsilon}(y_i,x_i,f(x_i)) = \nonumber \\ & \hspace{10mm} \begin{cases}
	\tau_2(|y_i-f(x_i)|-\epsilon),~ if~|y_i-f(x_i)|\geq \epsilon,\\
	\tau_1(|y_i-f(x_i)|-\epsilon) ~~~otherwise,
	\end{cases}
	\end{eqnarray} }

where $\tau_2 \geq \tau_1 $ and $\epsilon > 0$ are parameters. Figure \ref{:RewardPenalitylossfunction} shows the graph of  a typical reward cum penalty loss function for different values of $\tau_2 \geq \tau_1 \geq 0$. The proposed reward cum penalty loss function reduces to the popular $\epsilon$-insensitive loss function for $\tau_2 =1$ and $\tau_1 =0$. Figure \ref{:RewardPenalitylossfunctionwithnegativevalue} shows the graph of reward cum penalty loss for different values of $\tau_2 \geq \tau_1 \leq 0$.  It can also be observed that the proposed reward cum penalty loss function is a convex function for $\tau_2 \geq \tau_1 \geq 0$ but, for $\tau_1 \leq 0$, it loses its convexity. Therefore, in our subsequent discussion we shall always assume  $\tau_2 \geq \tau_1 \geq 0$.

To build the regression model based on the proposed reward cum penalty loss function, we use the same for measuring the empirical risk of the training set which is minimized in the proposed optimization problem along with the regularization term $\frac{1}{2} w^Tw$. We term the resulting regression model as `Reward cum Penalty loss function based $\epsilon$- Support Vector Regression(~RP-$\epsilon$-SVR)' model. Following are some salient features of the proposed reward cum penalty loss function and resulting RP-$\epsilon$-SVR model.

\begin{figure}
	\centering
	
	\subfloat[]{\includegraphics[width = 3.5in,height=1.4in]{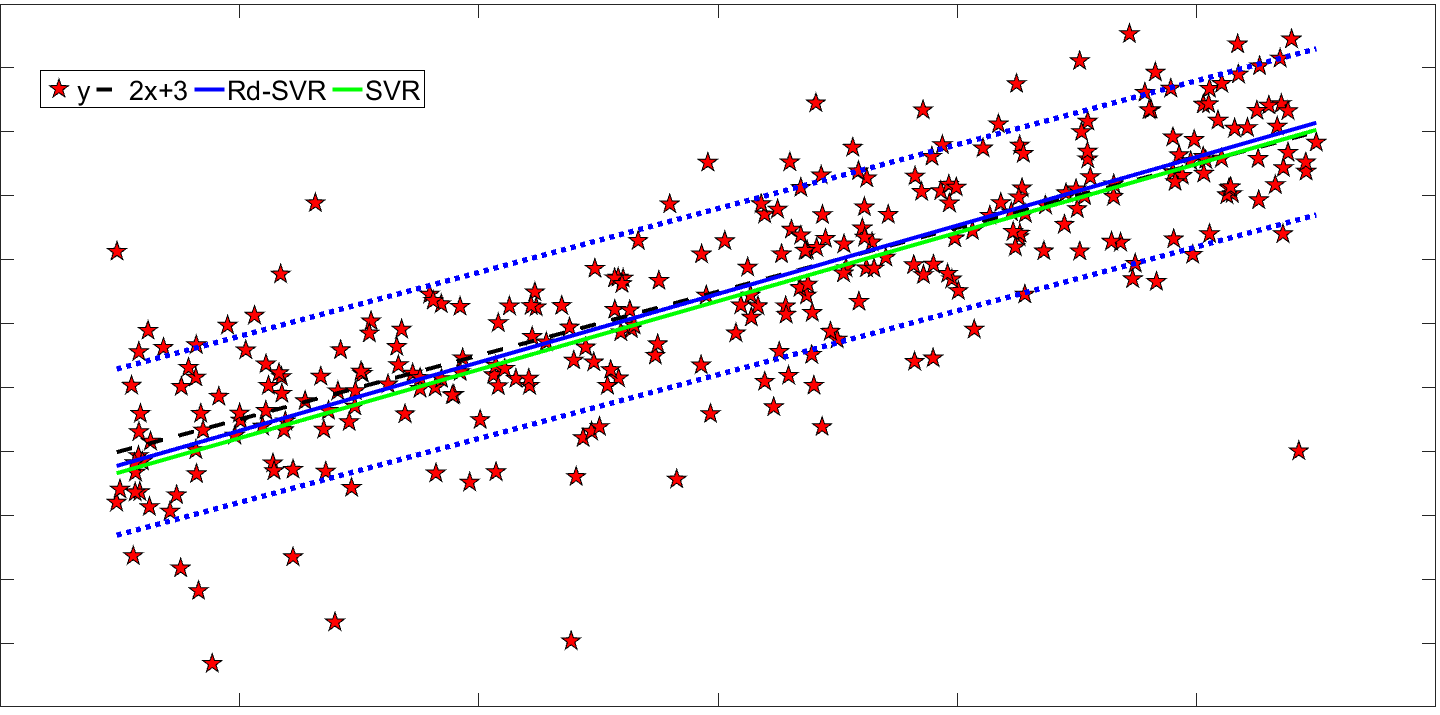}}\\
	\subfloat[]{\includegraphics[width = 3.5in,height=1.8in]{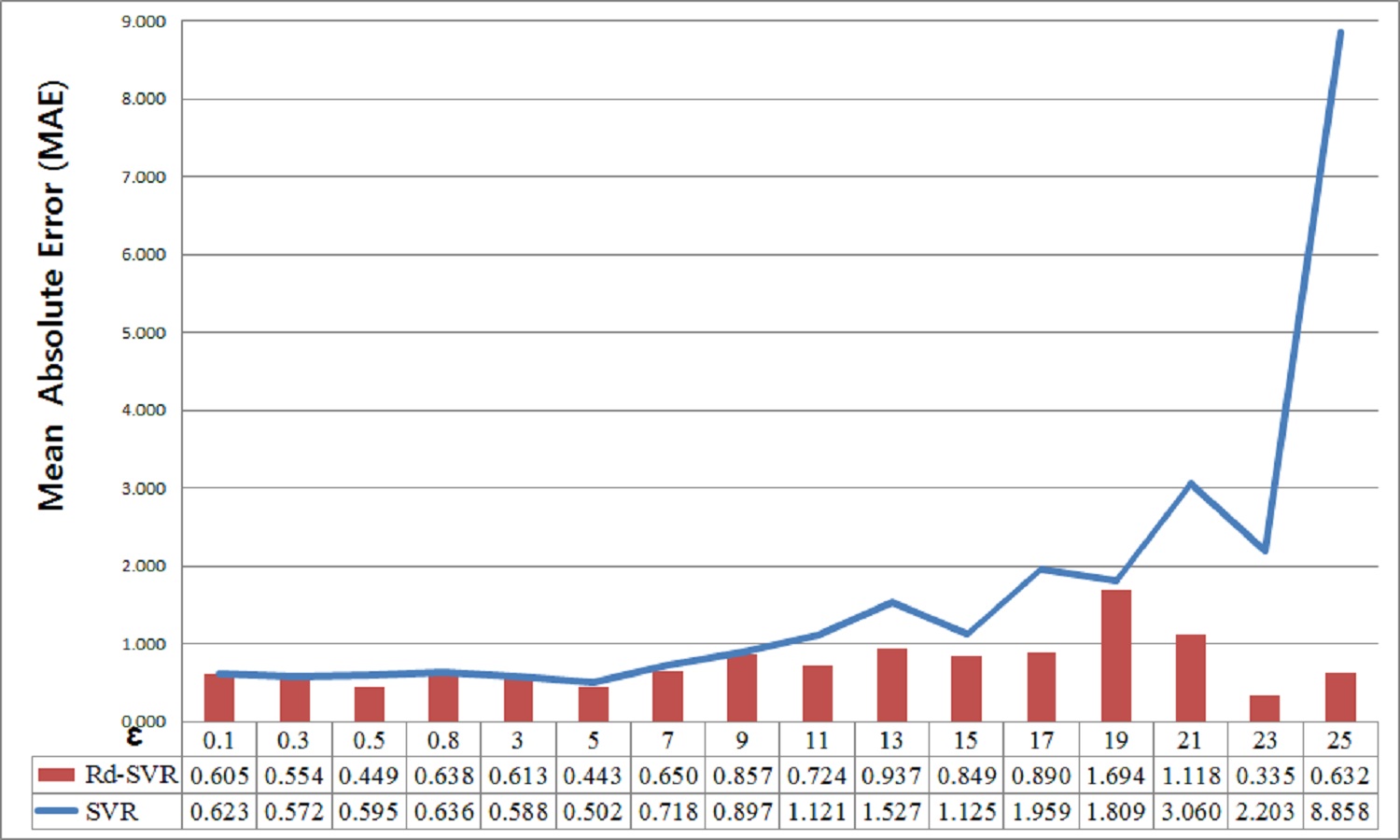}}
	
	\caption{ \scriptsize{We generate 600 data points form the U[ $-4\pi$ , $4\pi$]. The response $y =2x+ 3$ of the first $300$ training data points has been induced by the normal noise with mean 0 and variance = 10. After that, the response of the five training points has been polluted with the uniform noise from the U[-50,-25]  in order to introduce outliers. For testing, the response of the last 300 data points has not been polluted by any noise. (a)  One run simulation of the estimated function obtained by the RP-$\epsilon$-SVR model and $\epsilon$-SVR model on the  above mentioned artificial generated dataset is illustrated.
			(b) The proposed RP-$\epsilon$-SVR model almost always obtain lower RMSE than $\epsilon$-SVR model irrespective of the value of the $\epsilon$. Further as opposed to $\epsilon$-SVR, the performance of the RP-$\epsilon$-SVR is not much sensitive with the value of the $\epsilon$.}}  \
	\label{:perform_type1_dataset}
\end{figure}

\begin{enumerate}
	\item[(i)] The  reward cum penalty loss function is a  robust and convex loss function for $\tau_2 \geq \tau_1 \geq 0$.  It makes the optimization problem of the proposed RP-$\epsilon$-SVR formulation  a convex programing problem for $\tau_2 \geq \tau_1 \geq 0$ which can  therefore be solved efficiently. The robustness of the reward cum penalty loss function has been briefly established in Section \ref{robust} of this paper.
	\item[(ii)] The use of reward cum penalty loss function in SVR model can assign the penalty  $\tau_2(|y_i-f(x_i)|-\epsilon)$ for data points which lie outside the $\epsilon$-tube. The data points which lie inside the $\epsilon$-tube are assigned a reward $-\tau_1(|y_i-f(x_i)|-\epsilon)$.  The trade-off between the reward and  the penalty can be controlled by parameters $\tau_1$ and $\tau_2$. In this way, the proposed reward cum penalty loss function can properly use the full information of training set in a SVR model which was missing in the $\epsilon$-SVR model.
	\item [(iii)] We have judiciously used the proposed reward cum penalty loss function in proposed  RP-$\epsilon$-SVR model such a way that it can always obtain the sparse solution vector. The sparsity of the proposed RP-$\epsilon$-SVR model has also been  briefly derived through prepositions in Section \ref{sparsity} of this paper. The RP-$\epsilon$-SVR model can properly use the information of training set by associating a non-zero empirical risk with every training data point and can also preserve the robustness and sparsity property of the $\epsilon$-SVR model simultaneously. It makes the proposed RP-$\epsilon$-SVR model to obtain the better estimates always than $\epsilon$-SVR model always.
	
	\item[(iv)] The reward cum penalty loss function used in the RP-$\epsilon$-SVR model is a more general loss function. Section \ref{loss_density} of this paper shows that the use of reward cum penalty loss function in proposed RP-$\epsilon$-SVR model enables it to obtain the optimal estimate for a  large family of noise distribution. The well known Vapnik and Laplace noise distribution belong to this family of noise distribution with particular values of $\tau_2$ and $\tau_1$. That is why, the proposed RP-$\epsilon$-SVR model is expected to have better generalization ability than $\epsilon$-SVR model. 

	\item[(v)] In the $\epsilon$-SVR model, a good choice of the value of $\epsilon$ is required as the value of the $\epsilon$ decides that which of the  training points will be ignored and which of them will participate in the estimation of the regressor. That is  why, the performance of the $\epsilon$-SVR  model is much sensitive to the value of the $\epsilon$. But, the performance of the RP-$\epsilon$-SVR model is not much sensitive to its $\epsilon$ value. Though an extensive experimentation is reported in the experimental section of this paper, a simple MATLAB(in.mathworks.com) simulation in Fig. \ref{:perform_type1_dataset} shows the efficacy of the RP-$\epsilon$-SVR model over $\epsilon$-SVR model in this regard. It can be observed from Fig. \ref{:perform_type1_dataset} that though the proposed  RP-$\epsilon$-SVR model also requires the presumption of $\epsilon \geq 0$ but, it is not much sensitive to the value of the $\epsilon$ as it believes in the full utilization of the training set.      
	\item[(vi)]  The proposed RP-$\epsilon$-SVR model is based on the concept of the reward cum penalty loss function. Though in the best of our knowledge, there does not seem to exist any direct concept of this nature in the  regresssion literature  but, some of the works like (Huang et al. \cite{path}) use similar idea indirectly in the context of classification.
\end{enumerate}             

We now describe notations used in the rest of this paper. All vectors are taken as column vector unless it has been specified otherwise. For any vector $x \in R^{n}$, $||x||$  denotes the $L_2$ norm. A vector of ones of arbitrary dimension is denoted by $e$. $(A,Y)$ denotes the training set where $A = [A_1,A_2,.....,A_l]$ contains the $l$ points in $\mathbb{R}^n$ represented by $l$ rows of the matrix $A$ and $Y  = [y_1;y_2;...;y_l]\in \mathbb{R}^{l\times 1}$ contains the corresponding label or response value of the row of matrix $A$. Further, $\xi = (\xi^{1}; \xi^{2}; ..;\xi^{l})$, $\xi_1 = (\xi_1^{1}; \xi_1^{2}; ..;\xi_1^{l})$ and $\xi_2 = (\xi_2^{1}; \xi_2^{2}; ..; \xi_1^{l})$ are $l$ dimensional column vectors which will be used to denote the errors.

The rest of this paper has been organized as follows. Section \ref{sec2} briefly describes existing $\epsilon$- SVR model. In Section \ref{sec3}, the proposed RP-$\epsilon$-SVR model has been formulated for its linear and non-linear cases. In Section \ref{sec4}, we have  theoretically established  the robustness, sparsity and general nature of the proposed RP-$\epsilon$-SVR model. Section \ref{sec5} evaluates the proposed RP-$\epsilon$-SVR model using the numerical results which is obtained by the experiments carried on several artificial and UCI benchmark datasets. Section \ref{sec6} concludes this paper.                             
  
 \section {$\epsilon$- Support Vector Regression}
 \label{sec2}
 The standard $\epsilon$-SVR minimizes
 \begin{equation*}
 \frac{1}{2}||w||^2 + C\sum_{i=1}^{l}{L _{\epsilon}(y_i,x_i,f(x_i))}, 
 \end{equation*} 
 which can be equivalently converted to the following Quadratic Programming Problem (QPP)
 
 \begin{eqnarray}
 \min_{_{w,b,\xi_1 ,\xi_2}} ~ ~ \frac{1}{2}\Arrowvert w \Arrowvert^{2} + C e^T(\xi_1 + \xi_2)\nonumber\\
 & \hspace{-80mm }\hbox{subject to,} \nonumber \\
 & \hspace{-50mm }Y - (Aw + eb) \leq \epsilon e+ \xi_1, \nonumber\\
 & \hspace{-50mm }(Aw +eb) -Y \leq \epsilon e +\xi_2,\nonumber \\
 & \hspace{-60mm }\xi_1 \geq 0 , ~ \xi_2 \geq 0.
 \label{SvrPrimal}
 \end{eqnarray}
 Here $C >0$ is the user specified  positive parameter that balances the trade off between the training error and the flatness of the approximating function. To solve the primal problem (\ref{SvrPrimal}) efficiently, we write the corresponding Wolfe dual (Mangasarian, \cite{wolfedual}) using Karush-Kuhn-Tucker (KKT) conditions. The Wolfe dual of the primal problem (\ref{SvrPrimal}) has been obtained as follows.
 \begin{eqnarray}
  \min_{(\beta_1,\beta_2)} \frac{1}{2}(\beta_1-\beta_2)AA^{T}(\beta_1-\beta_2)-(\beta_1-\beta_2)^TY\nonumber\\
      & \hspace{-100mm} + (\beta_1+\beta_2)^{T}\epsilon e \nonumber\\
 & \hspace{-140mm}\mbox{subject to,} \nonumber \\
 & \hspace{-110mm}  (\beta_1 -\beta_2)^Te = 0, \nonumber\\
 &\hspace{-115mm} 0 \leq \beta_1, ~ \beta_2 \leq C . \label{dualsvr}
 \end{eqnarray}    
 After finding the optimal values of $\beta_1$ and  $\beta_2$, the estimated value for the test point $x$ is given by
 $f (x) =  (\beta_1-\beta_2)^TAx + b$.             
 
 \section{Reward cum Penalty $\epsilon$-Support Vector Regression }  
 \label{sec3}
 The RP-$\epsilon$-SVR model minimizes
 
 \begin{eqnarray}
 \frac{1}{2}||w||^2 + C\sum_{i=1}^{l}{RP _{(\tau_1,\tau_2,\epsilon)}(y_i,x_i,f(x_i))} 
 = \frac{1}{2}||w||^2+ \nonumber \\ C\sum_{i=1}^{l}\max(~\tau_2(|y_i-f(x_i)|-\epsilon), ~\tau_1(|y_i-f(x_i)|-\epsilon)),~ \label{mainpr2}
 \end{eqnarray} 
 where $\tau_2 \geq \tau_1 \geq 0$ and $\epsilon > 0$ are parameters.  Let us introduce a $l$-dimensional column error vector $\xi$ where $\xi^{i}$ = $\max(\tau_2(|y_i-f(x_i)|-\epsilon), ~\tau_1(|y_i-f(x_i)|-\epsilon))$ for $i=1,2,....l$. Then problem (\ref{mainpr2}) can be written as 
 follows 
 \begin{eqnarray}
 \min_{_{(w,b,\xi)}} ~ ~ \frac{1}{2}\Arrowvert w \Arrowvert^{2} + Ce^T\xi\nonumber\\
 & \hspace{-65mm }\hbox{subject to,} \nonumber \\
 & \hspace{-20mm }   \xi_i \geq  \tau_2(|y_i-f(x_i)|-\epsilon), ~~i=1,2,....l,  \nonumber\\
 & \hspace{-20mm }\xi_i \geq  \tau_1(|y_i-f(x_i)|-\epsilon),~~i=1,2,....l,
 \label{rdsvr1}
 \end{eqnarray}
 
 \subsection{ Linear Reward cum Penalty-$\epsilon$ SVR}
 The optimization problem (\ref{rdsvr1}) can be converted to the following standard QPP 
 \begin{eqnarray}
 \min_{_{(w,b,\xi_1 ,\xi_2)}} ~ ~ \frac{1}{2}\Arrowvert w \Arrowvert^{2} + Ce^T(\xi_1 + \xi_2)\nonumber\\
 & \hspace{-76mm }\hbox{subject to,} \nonumber \\
 & \hspace{-45mm } Y - (Aw +eb) \leq   \epsilon e + \frac{1}{\tau_1}\xi_1, \nonumber\\
 & \hspace{-45mm }(Aw +eb)-Y \leq \epsilon e + \frac{1}{\tau_1}\xi_2,\nonumber \\
 & \hspace{-42mm } Y - (Aw +eb) \leq \epsilon e + \frac{1}{\tau_2}\xi_1 ,\nonumber \\                                     
 & \hspace{-42mm } (Aw +eb)-Y \leq \epsilon e + \frac{1}{\tau_2}\xi_2,
 \label{rdsvr}
 \end{eqnarray}
 where $\xi_1$ and $\xi_2$ are $l$-dimensional slack variables.    
 The QPP (\ref{rdsvr}) reduces to QPP (\ref{SvrPrimal}) of the standard $\epsilon$-SVR model with the particular choice of parameters $\tau_2$ =1 and $\tau_1 =0$. 
 It makes the standard $\epsilon$-SVR model a particular case of the proposed RP-$\epsilon$-SVR formulation. 
 
 In order to find a solution of primal problem (\ref{rdsvr}), we need to derive its Wolfe dual (Mangasarian, \cite{wolfedual}). For this, we write the Lagrangian function for primal problem (\ref{rdsvr}) as follows\\
 $L(w,b,\xi_1,\xi_2,\alpha_1,\alpha_2,\beta_1,\beta_2) = \frac{1}{2}\Arrowvert w \Arrowvert^{2} + C e^T(\xi_1+\xi_2)+\alpha_1^{T}(Y -(Aw +eb)-\epsilon e- \frac{1}{\tau_1}\xi_1) +\alpha_2^{T}(Aw +eb-Y-\epsilon e-\frac{1}{\tau_1}\xi_2)+ \beta_1^{T}(Y-(Aw +eb)-\epsilon e-\frac{1}{\tau_2}\xi_1)+\beta_2^T(Aw +eb-Y-\epsilon e-\frac{1}{\tau_2}\xi_2),$\\
 where $\alpha_1 = {(\alpha_1^1,\alpha_1^2,....,\alpha_1^l)},~\alpha_2 = {(\alpha_2^1,\alpha_2^2,....,\alpha_2^l)},~\beta_1 = {(\beta_1^1,\beta_1^2,....,\beta_1^l)}$ and $\beta_2 ={(\beta_2^1,\beta_2^2,....,\beta_2^l)}$ are vectors of Lagrangian multipliers.\\
 The KKT optimality conditions  for the optimization problem (\ref{rdsvr}) are given by
 \begin{eqnarray}
 \frac{\partial L}{\partial w} =  w- A^T(\alpha_1- \alpha_2+ \beta_1 -\beta_2) = 0,\label{r11}\\
 &\hspace{-75mm} \frac{\partial L}{\partial b} =  e^T(\alpha_1- \alpha_2+ \beta_1 -\beta_2) = 0, \\
 & \hspace{-84mm}\frac{\partial L}{\partial \xi_1} =  C- \frac{1}{\tau_1}\alpha_1 -\frac{1}{\tau_2}\beta_1 = 0,\label{r1} \\
 & \hspace{-84mm} \frac{\partial L}{\partial \xi_2} =  C- \frac{1}{\tau_1}\alpha_2 -\frac{1}{\tau_2}\beta_2 = 0,\label{r4}\\
 &  \hspace{-72mm}\alpha_1^{T}(Y - (Aw +eb)- \epsilon e- \frac{1}{\tau_1}\xi_1) = 0 \label{r2},\\
 &  \hspace{-76mm} \alpha_2^{T}(Aw +eb-Y - \epsilon e-\frac{1}{\tau_1}\xi_2)=0 \label{r5},\\
 & \hspace{-75mm}  \beta_1^{T}(Y-(Aw +eb)-\epsilon e-\frac{1}{\tau_2}\xi_1)=0,\label{r3}\\                       
 & \hspace{-76mm}  \beta_2^T (Aw +eb-Y -\epsilon e- \frac{1}{\tau_2}\xi_2)=0,\label{r6} \\
 &  \hspace{-86mm} Y - (Aw +eb) \leq \epsilon e+ \frac{1}{\tau_1}\xi_1, \label{r7}\\                       
 & \hspace{-86mm}(Aw +eb)-Y \leq  \epsilon e + \frac{1}{\tau_1}\xi_2,\label{r8} \\
 & \hspace{-86mm} Y - (Aw +eb) \leq \epsilon e + \frac{1}{\tau_2}\xi_1 , \label{r9} \\  
 &  \hspace{-86mm}(Aw +eb)-Y \leq \epsilon e + \frac{1}{\tau_2}\xi_2, \label{r10}\\
 & \hspace{-88mm}\alpha_1 \geq 0 , \alpha_2 \geq 0 ,\beta_1 \geq 0 ,\beta_2 \geq 0.
 \end{eqnarray} 
 Using the above KKT conditions, the Wolfe dual (Mangasarian, \cite{wolfedual}) of primal problem (\ref{rdsvr}) can be obtained as follows        
 \begin{eqnarray}
 \min_{(\alpha_1,\alpha_2,\beta_1,\beta_2)} \frac{1}{2} (\alpha_1-\alpha_2+\beta_1-\beta_2)^TAA^{T}(\alpha_1-\alpha_2+\beta_1-\beta_2)\nonumber \\ - (\alpha_1-\alpha_2+\beta_1-\beta_2)^TY+ (\alpha_1+\alpha_2+\beta_1+\beta_2)^{T}e\epsilon  \nonumber\\
 & \hspace{-160mm}\mbox{subject to,} \nonumber \\
 & \hspace{-110mm}  (\alpha_1- \alpha_2+ \beta_1 -\beta_2)^Te = 0, \nonumber\\
 &\hspace{-115mm}  C- \frac{1}{\tau_1}\alpha_1 -\frac{1}{\tau_2}\beta_1=0,\nonumber \\
 & \hspace{-115mm}  C- \frac{1}{\tau_1}\alpha_2 -\frac{1}{\tau_2}\beta_2 = 0, \nonumber\\
 & \hspace{-120mm}  \alpha_1, \alpha_2, \beta_1,   \beta_2 \geq 0.\label{dualrdsvr}
 \end{eqnarray}    
 
 After obtaining the solution of the dual problem (\ref{dualrdsvr}), the value of $w$ can be obtained from the KKT condition (\ref{r11}) as follows
 \begin{eqnarray}
 w= A^T(\alpha_1- \alpha_2+\beta_1 -\beta_2).
 \end{eqnarray}         
 Let us now define the following sets\\
 $~~~~~~~~~~~~~~~~~ S_1= \{i:\alpha_1^{i} > 0 , \beta_1^{i} > 0   \}$,\\
 ~~~~~ and $ ~~~~~~~~~~~~S_2= \{j:\alpha_2^{j} > 0 , \beta_2^{j} > 0   \}$.  \\    \\ 
 Then taking  $i \in S_1$ and making use of the KKT conditions (\ref{r2}) and (\ref{r3}), we get
 \begin{eqnarray}
 & \hspace{-10mm}y_i - (A_iw +b)- \epsilon -\frac{1}{\tau_1}\xi_1^{i} =0, \label{rbb1}\\
 \mbox{and}~~~~~~~ & \hspace{-10mm}~~y_i - (A_iw +b)- \epsilon -\frac{1}{\tau_2}\xi_1^{i} =0\label{rbb2}. 
 \end{eqnarray}
 But (\ref{rbb1}) and (\ref{rbb2}) give $\xi_1^{i}(\frac{1}{\tau_2}-\frac{1}{\tau_1}) = 0$. Therefore for $\tau_1 \neq \tau_2$, we obtain 
 \begin{eqnarray}
 b = y_i-A_iw-\epsilon.  \label{rbcal1}
 \end{eqnarray} 
 
 On  similar lines,  taking$ ~~j\in S_2$ and $\tau_1 \neq \tau_2$ , we  obtain 
 \begin{eqnarray}
 b = y_j-A_jw+\epsilon. \label{rbcal2}
 \end{eqnarray}                                                
 In practice, for each $i \in S_1$ and each $j \in S_2$, we calculate the values of  $b$ from (\ref{rbcal1}) and (\ref{rbcal2}) respectively and take their average value as the final value of  $b$.
 For  the given test point $x \in {R}^n$,  the estimated response is obtained 
 \begin{eqnarray}
 f(x)  = w^Tx+ b = (\alpha_1-\alpha_2+\beta_1 -\beta_2)^TAx +b.
 \end{eqnarray}
 \subsection{Non-linear Reward cum Penalty-$\epsilon$ SVR}
 The non-linear RP-$\epsilon$-SVR model seeks to determine the  regressor\\
 $~~~~~~~f(x) = w^T\phi(x) +b$, where $\phi : R^n  \rightarrow \mathcal{H} $ is a non-linear mapping and $\mathcal{H}$ is  an appropriate higher dimensional feature space.
 
 The non-linear RP-$\epsilon$-SVR model solves the following optimization problem                            
 \begin{eqnarray}
 \min_{_{(w,b,\xi_1 ,\xi_2)}} ~ ~ \frac{1}{2}\Arrowvert w \Arrowvert^{2} + Ce^T(\xi_1 + \xi_2)\nonumber\\
 & \hspace{-76mm }\hbox{subject to,} \nonumber \\
 & \hspace{-45mm } Y - (\phi(A)w +eb) \leq   \epsilon e + \frac{1}{\tau_1}\xi_1, \nonumber\\
 & \hspace{-45mm }(\phi(A)w +eb)-Y \leq \epsilon e + \frac{1}{\tau_1}\xi_2,\nonumber \\
 & \hspace{-42mm } Y - (\phi(A)w +eb) \leq \epsilon e + \frac{1}{\tau_2}\xi_1 ,\nonumber \\                                     
 & \hspace{-42mm } (\phi(A)w +eb)-Y \leq \epsilon e + \frac{1}{\tau_2}\xi_2.
 \label{nlrdsvr}
 \end{eqnarray}              
 Similar to the linear RP-$\epsilon$-SVR  model, the corresponding Wolfe dual (Mangasarian, \cite{wolfedual}) problem of the primal problem (\ref{nlrdsvr}) is obtained as
 
 \begin{eqnarray}
 \min_{(\gamma_1,\gamma_2,\lambda_1,\lambda_2)} \frac{1}{2} (\gamma_1-\gamma_2+\lambda_1-\lambda_2)^T\phi(A)\phi(A)^{T}(\gamma_1-\gamma_2+\lambda_1-\lambda_2)\nonumber \\ - (\gamma_1-\gamma_2+\lambda_1-\lambda_2)^TY+ (\gamma_1+\gamma_2+\lambda_1+\lambda_2)^{T}e\epsilon  \nonumber\\
 & \hspace{-160mm}\mbox{subject to,} \nonumber \\
 & \hspace{-110mm}  (\gamma_1- \gamma_2+ \lambda_1 -\lambda_2)^Te = 0, \nonumber\\
 &\hspace{-115mm}  C- \frac{1}{\tau_1}\gamma_1 -\frac{1}{\tau_2}\lambda_1=0,\nonumber \\
 & \hspace{-115mm}  C- \frac{1}{\tau_1}\gamma_2 -\frac{1}{\tau_2}\lambda_2 = 0 , \nonumber \\
 & \hspace{-115mm}  \gamma_1, \gamma_2, \lambda_1,   \lambda_2 \geq 0.                                 \label{dualrdsvrnonlin}
 \end{eqnarray}    
 A positive definite kernel $K(A,A^T)$, satisfying the Mercer condition (Scholkopf and Smola  \cite{kernel}), is used to obtain $\phi(A)\phi(A)^T$ without explicit knowledge of mapping $\phi$. Thus problem (\ref{dualrdsvrnonlin}) reduces to                                                                      
 \begin{eqnarray}
 \min_{(\gamma_1,\gamma_2,\lambda_1,\lambda_2)} \frac{1}{2} (\gamma_1-\gamma_2+\lambda_1-\lambda_2)^TK(A,A^T)(\gamma_1-\gamma_2+\lambda_1-\lambda_2)\nonumber \\ 
 &\hspace{-95mm} - (\gamma_1-\gamma_2+\lambda_1-\lambda_2)^TY+ (\gamma_1+\gamma_2+\lambda_1+\lambda_2)^{T}\epsilon e \nonumber\\
 & \hspace{-160mm}\mbox{subject to,} \nonumber \\
 & \hspace{-110mm}  (\gamma_1- \gamma_2+ \lambda_1 -\lambda_2)^Te = 0, \nonumber\\
 &\hspace{-115mm}  C- \frac{1}{\tau_1}\gamma_1 -\frac{1}{\tau_2}\lambda_1=0,\nonumber \\
 & \hspace{-115mm}  \gamma_1, \gamma_2, \lambda_1,   \lambda_2 \geq 0.                                 
 \end{eqnarray}    
 
 For the given test point $x \in {R}^n$,  the determined regressor gives the value
 \begin{eqnarray}
 f(x)  = w^T\phi(x) +b  \nonumber\\
 & \hspace{-30mm}=  (\gamma_1- \gamma_2+\lambda_1 -\lambda_2)^TK(A,x) + b.
 \end{eqnarray}

 \section{Properties of proposed RP-$\epsilon$-SVR model }
 \label{sec4}
 
 \subsection{Maximal likelihood approach and loss functions}\label{loss_density}
 Let $T= \{ (x_i,~y_i),~x_i \in \mathbb{R}^n,~ y_i \in \mathbb{R},~i~=~1,~2,~...l \}$ be the given training set.
 It is assumed that values $(x_i,y_i)$ are related by unknown function $f$ such that 
 \begin{equation}
 y_i = f(x_i)+ \xi_i,
 \end{equation}
 where $\xi_i$ are independent and identically distributed random variables form an unknown distribution $p(\xi)$. The celebrated Statistical Learning Theory  (Vapnik, \cite{statistical_learning_theory}) employs the maximal likelihood principle to derive the `optimal' loss function for a given distribution function  $p(\xi)$. This `optimal' loss function is used to determine the regressor $f$ for the estimation of the response $y_j$ for a given test data point $x_j$. Here the `optimal' is understood in terms of maximizing the 'likelihood function' for the given training set $T$, which is given by
 \begin{equation}
 p[T/f] = \prod_{i=1}^{l} p(y_i-f(x_i)) = \prod_{i=1}^{l} p(\xi_i).\label{likelyfunc}
 \end{equation}
 Since  $ p(\xi_i) \geq 0 $ for all $i$, the maximization of the likelihood function (\ref{likelyfunc}) is equivalent to the maximization of the log of the likelihood function. Therefore (\ref{likelyfunc}) is equivalent to
 \begin{equation}
 min~~ \sum_{i=1}^{l} -log(p(\xi_i)). \label{cc}
 \end{equation}
 Now the specific assumption about the density of noise model will specify the computed loss function which should be used for measuring the empirical error for finding the estimator function $f$. We describe following robust densities of noise which lead to different popular loss functions.
 \begin{enumerate}
 	\item[(i)]
 	Laplace noise distribution: \\
 	This noise model is given by
 	\begin{equation}
 	p(\xi) \propto \frac{1}{2} e ^{-|\xi|},~~\xi \in \mathbb{R}.\label{laplace}
 	\end{equation}
 	On substituting (\ref{laplace}) into (\ref{cc}), we get
 	\begin{figure}
 		\centering
 		\includegraphics[width=1.0\linewidth, height=0.15\textheight]{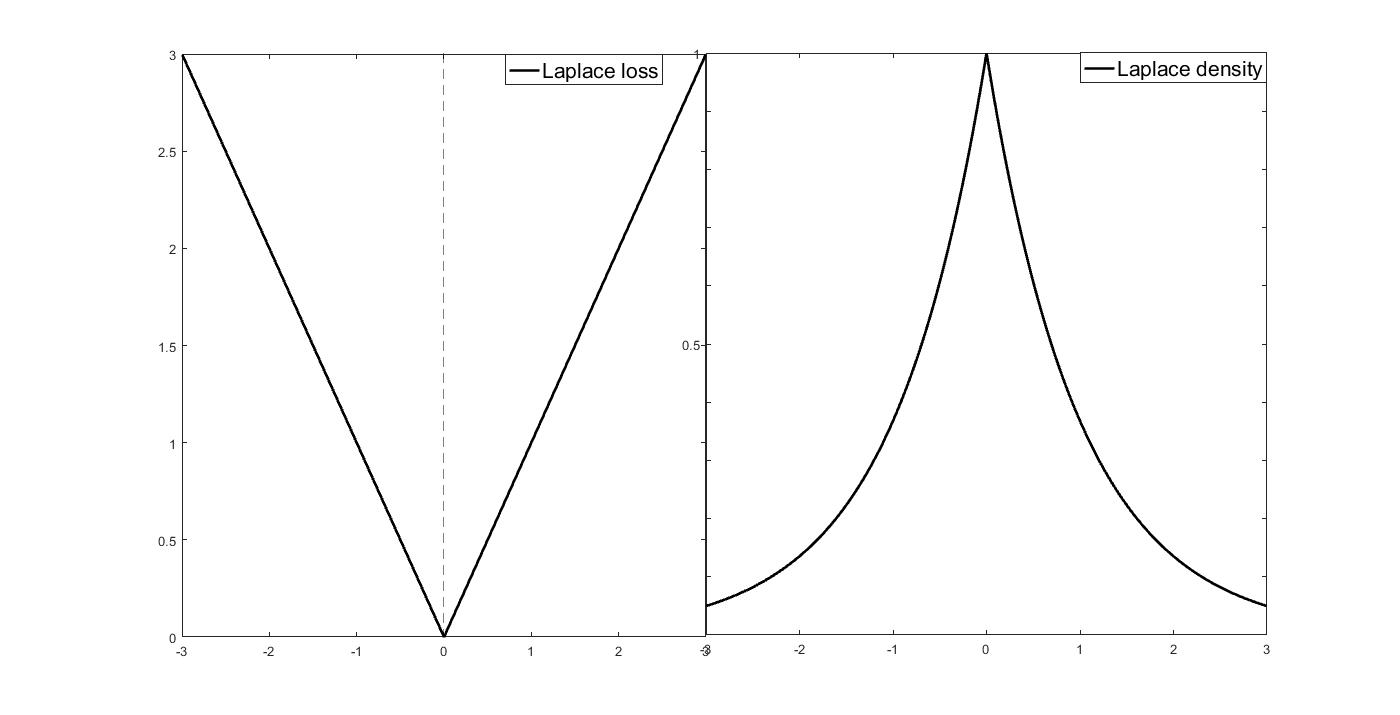}
 		\caption{Laplace loss function and corresponding density function.}
 		\label{laplace12}
 	\end{figure}
 	\begin{equation}
 	min~~ \sum_{i=1}^{l}|\xi_i|,
 	\end{equation} which is equivalent to the minimization of the Laplace loss  $L(\xi)  = |\xi|$ for the training set $T$. Fig \ref{laplace12} shows the Laplace loss function and its corresponding density function $p(\xi)$.

 	\item[(ii)] Vapnik distribution : \\
 	It is one of the popular noise models used in the standard SVR formulation and is defined as
 	\begin{equation}
 	p(\xi) \propto  \frac{1}{2(1+\epsilon)}e^{-|\xi|_\epsilon} \label{epsinsen}.
 	\end{equation}
 	On substituting (\ref{epsinsen}) into (\ref{cc}), we get
 	\begin{equation}
 	min~~ \sum_{i=1}^{l}|\xi_i|_\epsilon ,
 	\end{equation}
 	where~~~
 	\begin{equation}
 	|\xi_i|_\epsilon ~=~\begin{cases}
 	0, ~~~~~~~~~~~~ |\xi_i| <  \epsilon, \\
 	|\xi_i| -\epsilon, ~~~~~ otherwise, 
 	\end{cases} \nonumber
 	\end{equation}
 	is the $\epsilon$-insensitive loss function used in the standard SVR formulation. Fig \ref{epsilon12} shows the $\epsilon$-insensitive loss function and its corresponding density function.
 	\begin{figure}
 		\centering
 		\includegraphics[width=1.0\linewidth, height=0.15\textheight]{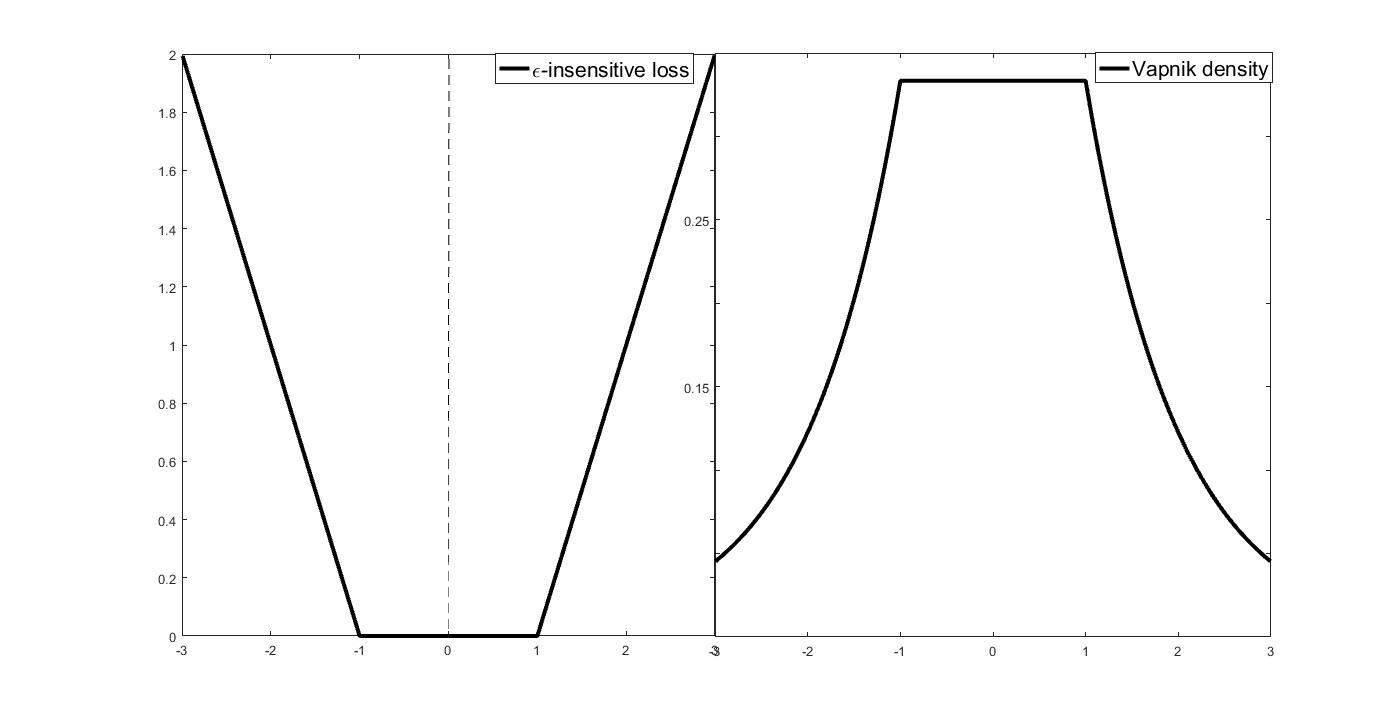}
 		\caption{$\epsilon$-insensitive loss function and corresponding density function.}
 		\label{epsilon12}
 	\end{figure}
 	
 	\begin{figure}
 		\centering
 		\includegraphics[width=1.0\linewidth, height=0.15\textheight]{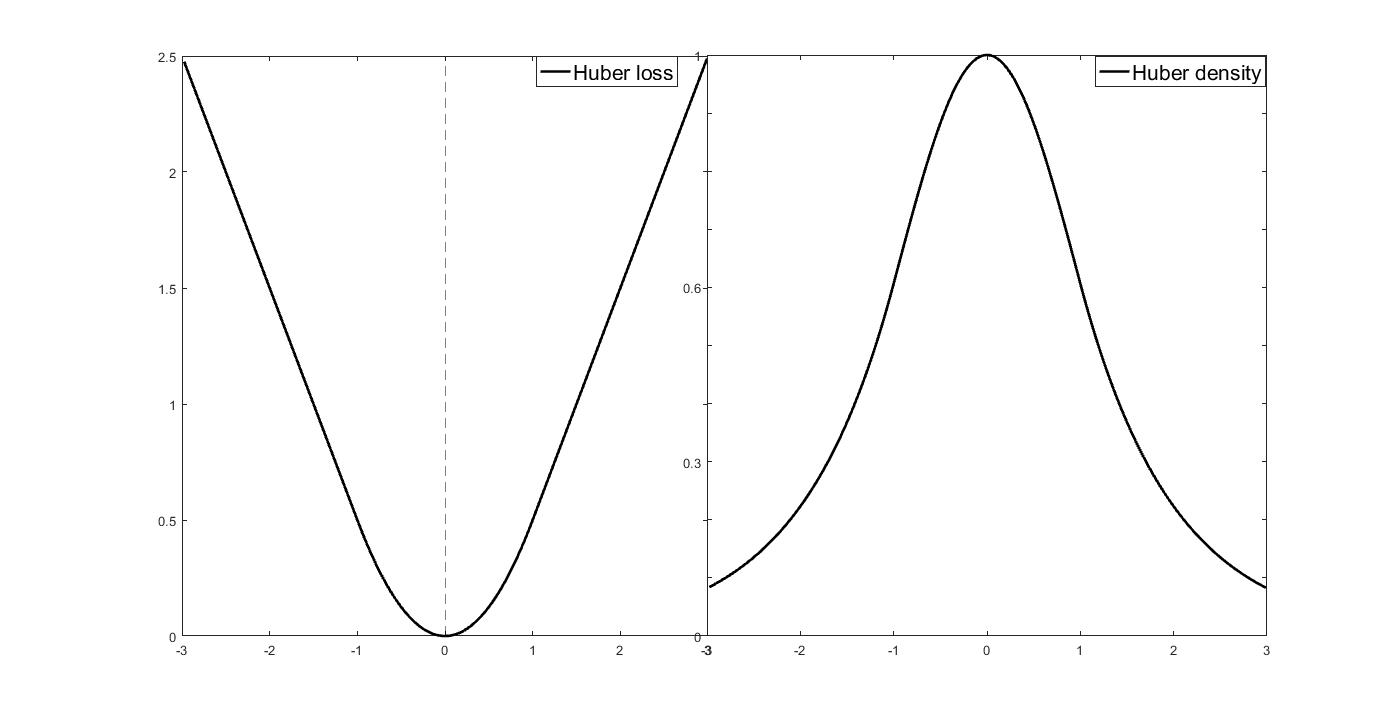}
 		\caption{Huber loss function and corresponding density function.}
 		\label{huber12}
 	\end{figure}
 	
 	\item[(iii)] Huber distribution :\\
 	It is a mixed noise model which is described as 
 	\begin{equation}
 	p(\xi) \propto  \begin{cases} e^{-\frac{1}{2c}\xi_i^2 },~~~ if ~~~|\xi| < c. \\
 	e^{(\frac{c}{2}-|\xi_i|)} , ~~ otherwise. \label{huberdensity}
 	\end{cases} 
 	\end{equation} 
 	On substituting (\ref{huberdensity}) in (\ref{cc}), we get
 	\begin{equation}
 	\min ~~L_{\epsilon} ^{Huber}
 	\end{equation}
 	where 
 	\begin{equation}
 	L_{\epsilon} ^{Huber}=
 	\begin{cases}
 	\sum_{i=1}^{l} \frac{1}{2c}(\xi_i)^2,~~~ if ~~~|\xi| < c, \\
 	\sum_{i=1}^{l}(|\xi_i| - \frac{c}{2}), ~~ otherwise,
 	\end{cases}
 	\end{equation}
 	is Huber loss function . Fig \ref{huber12} shows the Huber loss function and its corresponding density function.\\
 	\item[(iv)] Distribution of noise for the proposed reward cum penalty loss function: \\
 	We now present an analysis of above nature for our proposed reward cum penalty loss function $RP_{\tau_1,\tau_2}(u)$.
 	Let us consider a noise model which follows the density function 
 	\begin{equation}
 	p(\xi) \propto \frac{1}{2((\tau_2-\tau_1) + \epsilon)} ~e^{- \max(~\tau_2(|u|-\epsilon), ~\tau_1(|u|-\epsilon)~)}. \label{propseddensity}
 	\end{equation}
 	Substituting (\ref{propseddensity}) in (\ref{cc}) we get
 	\begin{equation}
 	\min~~  \sum_{i=1}^{l} \max(~\tau_2(|u|-\epsilon), ~\tau_1(|u|-\epsilon)~), \label{propsedlossfunction}
 	\end{equation} 
 	\begin{figure}
 		\centering
 		\includegraphics[width=1.1\linewidth, height=0.18\textheight]{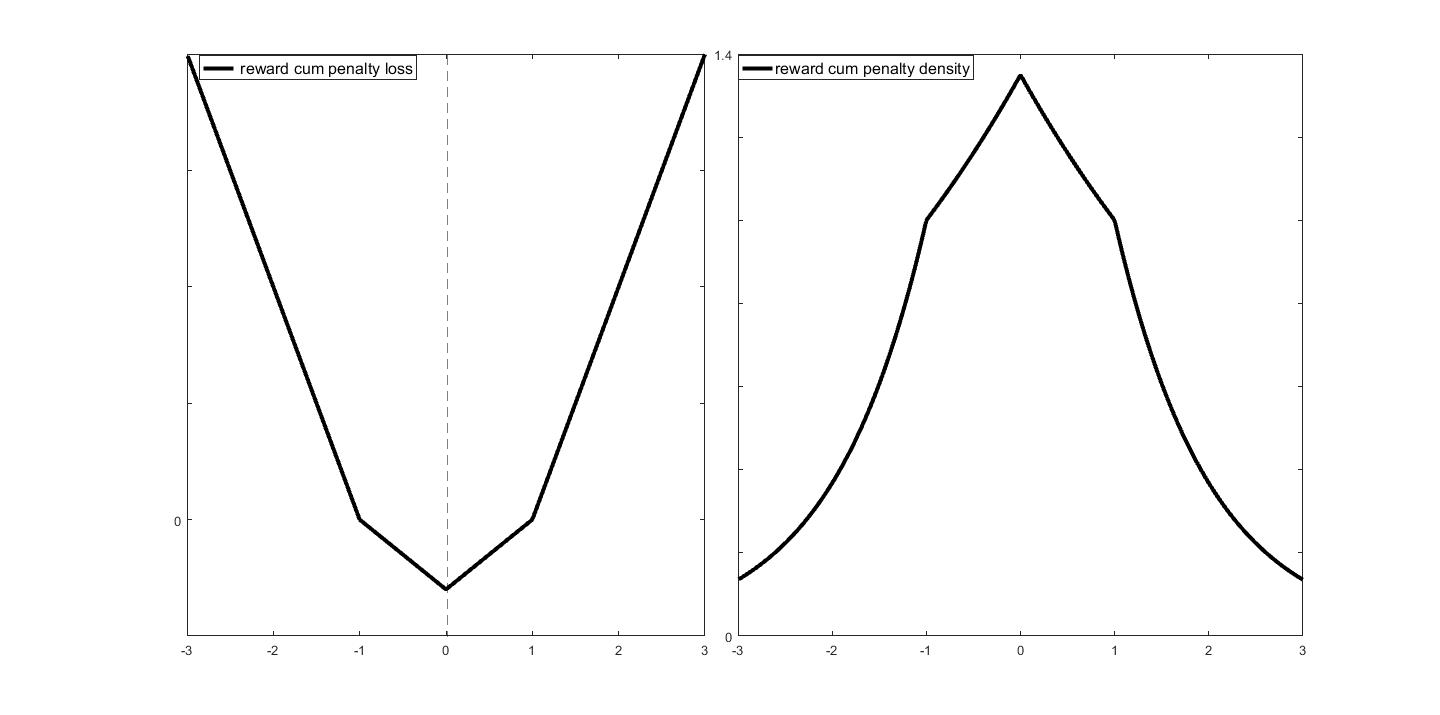}
 		\caption{Reward cum Penalty loss function and corresponding density function.}
 		\label{rewardcumpeanltydensity}
 	\end{figure}
 	
 	which is equivalent to the minimization of the proposed reward cum penalty loss function.  Fig \ref{rewardcumpeanltydensity} shows the proposed reward cum penalty loss function and its corresponding density function.
 	
 	Here, it is interesting to note that equation(\ref{propseddensity}) represents a family of noise densities for different choices of $\tau_1$ and $\tau_2$. Therefore, as a consequence, (\ref{propsedlossfunction}) represents a family of loss function for different value of $\tau_1$ and $\tau_2$. In particular, the density function of  Laplace distribution and Vapnik distribution belongs to the family of densities (\ref{propseddensity}) with the particular choice of the parameters $(\tau_2=1 ,\tau_1 =0)$ and  $(\tau_2=1 ,\tau_1 =1)$  respectively.   Hence, we can argue that the proposed loss function is a more general loss function in the sense that it is optimal to a wide range of  noise models which also include the Vapnik and Laplace noise models.   
 \end{enumerate}    
 
 \subsection{Sparsity of proposed RP-$\epsilon$-SVR model}
 \label{sparsity}
 \textbf{Preposition-1}
 For a given $\tau_2 >\tau_1$ and data point $(x_i,y_i)$, the  $\alpha_1^{i}\beta_1^{i}\neq 0$ or $\alpha_2^{i}\beta_2^{i} \neq 0 $ is possible, only when it is lying on the boundary of the $\epsilon$-tube.

 \textbf{Proof}:-  Let us consider first that $\alpha_1^{_i}\beta_1^{i} \neq 0$. It is possible only when $\alpha_1^{_i}$ and $\beta_1^{i} > 0$.  For $\alpha_1^{_i}$ and $\beta_1^{i} > 0$, we can obtain as follows
 \begin{eqnarray}                     (Y_i - (A_iw +b)- \epsilon - \frac{1}{\tau_1}\xi_1^{i}) = 0 \label{eqn1} \\
 \mbox{and}  ~~ (Y_i - (A_iw +b)- \epsilon - \frac{1}{\tau_2}\xi_1^{i}) = 0  \label{eqn2}      
 \end{eqnarray}
 from KKT condition (\ref{r2}) and (\ref{r3}) respectively.  After solving the equation (\ref{eqn1}) and (\ref{eqn2}) , we get $\xi_1^{i} =0$ as $\tau_2 \neq \tau_1$. It implies that $Y_i - (A_iw +b) = \epsilon$ which means that the response point $y_i$ for data point $(x_i,y_i)$ is lying on the  upper boundary of the $\epsilon$-tube.
 
 On the similar line, we can consider $\alpha_2^{_i}\beta_2^{i} \neq 0$ and can obtain $\xi_2^{i} =0$ from the KKT condition (\ref{r5}) and (\ref{r6}). It means that the response point $y_i$ for data point $(x_i,y_i)$ is lying on the lower boundary of the $\epsilon$-tube.

 The contra-positive statement equivalent to the Preposition-1 is as follow.
 \textit{ For any data point $(x_i,y_i)$, which is not lying on the boundary of the $\epsilon$-tube, i,e. lying inside or outside of the $\epsilon$-tube, the  $\alpha_1^{i}\beta_1^{i} =0 $ and $\alpha_2^{i}\beta_2^{i} = 0 $ will hold true.}

 \textbf{Preposition 2} For a given $\tau_2 >\tau_1$, any data point $(x_i,y_i)$ lying inside of the $\epsilon$-tube must satisfy $\alpha_1^{i}\beta_2^{i} =0 $ and $\alpha_2^{i}\beta_1^{i} =0 $.
 
 \textbf{Proof} Since data point $(x_i,y_i)$ is lying inside of the $\epsilon$-tube, so it will satisfy
 \begin{eqnarray}
 Y_i-(A_iw+b) -\epsilon  < 0  \\
 \mbox{and}~~ (A_iw+b)-Y_i- \epsilon  < 0.
 \end{eqnarray}    
 If possible, let us suppose that $\alpha_1^{i}\beta_2^{i} =0 $. It means that $\alpha_1^{i} > 0$ and  $\beta_2^{i} > 0$ from which we can obtain 
 \begin{eqnarray}
 \xi_1^{i} =\tau_1(Y_i - (A_iw +b)- \epsilon) \label{eq4} \\
 \mbox{and}~~\xi_2^{i} =\tau_2((A_iw +b)-Y_i-\epsilon). \label{eq5}
 \end{eqnarray}
 But, the KKT conditions (\ref{r8}) is
 \begin{equation}
 (A_iw +b)-Y_i \leq \epsilon  + \frac{1}{\tau_1}\xi_2^i.
 \end{equation}
 After putting the value of the $\xi_2^i$ from (\ref{eq5}) , we get  
 \begin{equation}
 (A_iw +b)-Y_i -\epsilon \leq \frac{\tau_2}{\tau_1}((A_iw +b)-Y_i-\epsilon)
 \end{equation}
 which is not possible as  $(A_iw +b)-Y_i -\epsilon < 0 $ and $\tau_2 > \tau_1$.  
 
 On the similar line, we can show that $\alpha_2^{i}\beta_1^{i} =0$ as $\alpha_2^{i} > 0$ and  $\beta_1^{i} > 0$  contradicts the KKT condition (\ref{r7}).        
 \\
 
 \textbf{Preposition 3} For $\tau_2 > \tau_1$, all data points $(x_i,y_i)$, which lie inside of the $\epsilon$-tube,  must satisfy $(\alpha_1^{i}- \alpha_2^{i}+\beta_1^{i} -\beta_2^{i}) =0$. \\    \textbf{Proof}:-  
 From the KKT condition (\ref{r1}) and (\ref{r4}), we can obtain  
 \begin{equation}
 \frac{1}{\tau_1}\alpha_1 + \frac{1}{\tau_2}\beta_1 = \frac{1}{\tau_1}\alpha_2 +\frac{1}{\tau_2}\beta_2 =C \label{eq22}
 \end{equation}
 Also, from Preposition-1 and Preposition-2 , we have 
 \begin{eqnarray}
 \alpha_1^{i}\beta_1^{i} = 0 , ~~\alpha_2^{i}\beta_2^{i} = 0. \\
 \mbox{and}~~ \alpha_1^{i}\beta_2^{i} = 0 ~~, \alpha_2^{i}\beta_1^{i} = 0.
 \end{eqnarray}
 respectively.
 From which, we can infer that there will exist only one of possible following cases when a data point $(x_i,y_i)$ is lying inside of the $\epsilon$-tube for a given $\tau_2 > \tau_1$.
 \begin{enumerate}
 	\item [(a)]  Only $\alpha_1^i$ and $\alpha_2^{i}$ takes non-zero values.
 	\item [(b)]  Only $\beta_1^i$ and $\beta_2^{i}$ takes non-zero values.  
 \end{enumerate} 
 
 But, in all of cases, we can get $(\alpha_1^{i}- \alpha_2^{i}+\beta_1^{i} -\beta_2^{i} =0)$  from (\ref{eq22}). It completes the proof.
 
 Though, the proposed RP-$\epsilon$-SVR model assigns a non-zero empirical risk with every training data point but, it can still obtain the sparse solution as we can obtain $(\alpha_1^{i}- \alpha_2^{i}+\beta_1^{i} -\beta_2^{i} =0)$ for all training data points which lie inside of the $\epsilon$-tube.

 \subsection{Robustness of the proposed $\epsilon$-penalty loss function}
 \label{robust}
 In this subsection, we shall show the robustness of proposed $\epsilon$-penalty loss function against outliers. For this, we shall be using the approach based on the influence function. This approach has been used to measure the robustness of loss functions in (Karal, \cite{incoshloss}). It has also been shown in (Karal, \cite{incoshloss}) that the influence function for the quadratic loss function is not bounded.  However, the influence function of the popular $\epsilon$-insensitive loss function is bounded. A loss function which has bounded influence function is a desirable loss function for a regression model as it makes the regression model a robust regression model.
 
 For a given training set, the loss function $L(u)$ is used to measure the empirical risk as follows.
 \begin{equation}
 E= \frac{1}{N}\sum_{i=1}^{N}L(u_i) \label{inf1}
 \end{equation}
 
 Taking the gradient of (\ref{inf1}) with respect to the model parameter $W$ will give
 \begin{equation}
 \frac{\partial E}{\partial W} = \frac{1}{N}\sum_{i=1}^{N}\bar{\phi}(u_i) \frac{\partial u_i}{\partial W} \label{inf2}
 \end{equation}
 Here $\bar{\phi}(u) =  \frac{\partial L(u)}{\partial u}$ is the influence function of loss function $L(u)$ which determines the performance of the loss function. We can observe from (\ref{inf2}) that the rate of the change in the empirical risk is the weighted sum of $\frac{\partial u_i}{\partial W}$ and $\bar{\phi}(u_i)$ stands for the weights for the data points. 
 
 The influence function of the proposed reward cum penalty loss function is given as follow
 \begin{eqnarray}
 \bar{\phi}(u_i) = \begin{cases}
 \tau_1 sign (u_i) ,~~~~~~~~~~~~~~~~~for ~~|u_i| \leq \epsilon , \\
 \tau_2 sign (u_i), ~~~~~~~~~~~~~~~~~~ for~~  |u_i| > \epsilon,
 \end{cases}
 \end{eqnarray}
 where \begin{equation}
 sign(u) =\begin{cases}
 -1, ~~~~~ for ~~~ u \leq  0 .  \\
 1, ~~~~~ for ~~~ u > 0 . \\
 \end{cases}
 \end{equation} 
 It means that, the influence function of the proposed $\epsilon$-penalty loss function is bounded in the interval $[ -\tau_2,\tau_2]$ and the effect of any sample point is limited in the range $[ -\tau_2,\tau_2]$.

 \section{Experimental Results}
 \label{sec5}
 To study the behavior of the proposed RP-$\epsilon$-SVR model, we have tested it on  eight artificial and  ten real world UCI benchmark (Blake CI and Merz CJ \cite{UCIbenchmark}) datasets. The proposed RP-$\epsilon$-SVR is basically  an improvement over the standard $\epsilon$-SVR model. Therefore, we have also compared the performance of the RP-$\epsilon$-SVR model with existing $\epsilon$-SVR model on these datasets. The numerical results on these datasets illustrate that irrespective of the nature of the noise present in these datasets, the proposed RP-$\epsilon$-SVR model always obtains better generalization ability than existing $\epsilon$-SVR model.
 
 All regression methods presented here were simulated in MATLAB 16.0 environment (http://in.mathworks.com/) on Intel XEON processor with 16.0 GB RAM. The  respective primal problems of the proposed RP-$\epsilon$-SVR and existing $\epsilon$-SVR models have same number of constraints and variables. However, the dual problem of the  proposed RP-$\epsilon$-SVR model has $4l$ variables, $2l+1$ equality constraints and $4l$ inequality constraints, where as the dual problem of the $\epsilon$-SVR model has $2l$ variables, $1$ equality constraints and $2l$ inequality constraints. The dual QPPs of the proposed RP-$\epsilon$-SVR model and $\epsilon$-SVR model have been solved by using the `quadprog' function of MATLAB (http://in.mathworks.com/) with its default algorithm in this paper. The development of an efficient algorithm for the solution of the QPP of the proposed RP-$\epsilon$-SVR model has been left as future work. Throughout the experiments, we have used RBF kernel $exp(\frac{-||x-y||^2}{q})$ where $q$ is the kernel parameter.
 
 The optimal values of the parameters have been obtained using the exhaustive search method (Hsu and Lin \cite{Exhaustivesearch}) by using cross-validation. The values of the parameter $C$ and RBF kernel parameter $q$ of $\epsilon$-SVR model have been tunned by searching in the set $\{2^i,~ i = -10,-2,....,12 \}$.  The value of the parameter $\epsilon$ of $\epsilon$-SVR model has been searched in the set $\{ 0.05,0.1,0.2,0.3 . . . . . ,1, 1.5 ,2,2.5,3,3.5,4,4.5,5\}$. To have a fair comparisons with $\epsilon$-SVR model, we have not explicitly tunned the parameters $q$, $C$ and $\epsilon$ of the RP-$\epsilon$-SVR model rather, we have used the same values of these parameters which was obtained form the $\epsilon$-SVR model. We have only  tunned the value of  parameters $\tau_1$ and $\tau_2$ of the proposed RP-$\epsilon$-SVR model by searching in the set $\{0.5, 0.6,. . . . . ,2.5\}$  and $\{0.1,0.2,0.3 . . . . 1\}$ respectively.
 \subsection{\textbf{Performance Criteria}}
 For evaluating the performance of the regression methods, we introduce some commonly used evaluation criteria. Without loss of generality, let $l$ and $k$ be the number of the training samples and testing samples respectively. Furthermore, for $i=1,2,...k$, let ${y'_i} $ be the predicted value for the response value $y_i$ and $\bar{y}$ = $\frac{1}{k}\sum_{i}^{k}y_i$ is the average of  $y_1, y_2,.....,y_k$. The definition and significance of the some evaluation criteria has been listed as follows.
 \begin{enumerate}
 	\item[(i)] SSE: Sum of squared error of testing, which is defined as SSE=$\sum_{i=1}^{k}(y_i-y'_i)^2$. SSE represents the fitting precision.
 	\item[(ii)] SST : Sum of squared deviation of testing samples, which is defined as SST = $\sum_{i=1}^{k}(y_i-\overline{y})^2$. SST shows the underlying variance of the testing samples.
 	\item[(iii)] SSR : Sum of square deviation of the testing samples which can be explained by the estimated regressor. It is defined as SSR =  $\sum_{i=1}^{k}(y'_i-\overline{y})^2$.
 	\item[(iv)] RMSE : Root mean square of  the testing error, which is defined as RMSE = $\sqrt{\frac{1}{k}\sum_{i=1}^{k}(y_i-y'_i)^2}$.
 	\item[(v)] MAE: Mean absolute error of testing, which is defined as $\frac{1}{k} \sum_{i=1}^{k}|(y_i-y'_i)|$.  
 	\item[(vi)] SSE/SST : SSE/SST is the ratio between the sum of the square of the testing error and sum of the square of the deviation of testing samples. In most cases, small SSE/SST means good agreement between estimations and real values.   
 	\item[(vii)] SSR/SST : It is the ratio between the variance obtained by the estimated regressor on testing samples and actual underlying variance of the testing samples. 
 	\item[(viii)]   Sparsity $\%$ : The sparsity of a vector $u$ is defined as \\ Sparsity $\%$ (u) = $\frac{\#(u=0)}{\#(u)}\times 100$, where $\#(r)$ determines the number of the component of the vector $r$.
 \end{enumerate} 
 
 \subsection{Artificial Datasets}\label{ad}
 We have synthesized some artificial datasets to show the efficacy of the  proposed method over the other existing methods . To compare the noise-insensitivty of the regression methods, only training sets were added with different types of noises in these artificial datasets.  For the  training samples $(x_i,y_i)$ for $i =1 ,2 ,..,l $,  following types of datasets have been generated.\\
 
 \textbf{TYPE 1:-}
 \begin{eqnarray*}
 	y_i = \frac{sin(x_i)}{x_i} + \xi_i ,~~\xi_i\sim U[-0.2,0.2]
 \end{eqnarray*}
 ~~~~~~~~~~ and  $x_i$ is from $U[-4\pi ,4\pi]$. 
 \\

 \textbf{TYPE 2:-}
 \begin{eqnarray*}
 	y_i = \frac{sin(x_i)}{x_i} + \xi_i, ~~\xi_i\sim U[-0.3,0.3]
 \end{eqnarray*}
 ~~~~~~~~~ and  $x_i$ is from $U[-4\pi ,4\pi]$. 
 \\
 
 \textbf{TYPE 3:-}
 \begin{eqnarray*}
 	y_i = \frac{sin(x_i)}{x_i} + \xi_i, ~~\xi_i\sim U[-0.4,0.4]
 \end{eqnarray*}
 ~~~~~~~~~ and  $x_i$ is from $U[-4\pi ,4\pi]$. 
 \\
 
 \textbf{TYPE 4:-}
 \begin{eqnarray*}
 	y_i = \frac{sin(x_i)}{x_i} + \xi_i, ~~\xi_i\sim N[0,0.1]
 \end{eqnarray*}
 ~~~~~~~~~~ and  $x_i$ is from $U[-4\pi ,4\pi]$. 
 \\

 \textbf{TYPE 5:-}
 \begin{eqnarray*}
 	y_i = \frac{sin(x_i)}{x_i} + \xi_i, ~~\xi_i\sim N[0,0.3]
 \end{eqnarray*}
 ~~~~~~~~~ and  $x_i$ is from $U[-4\pi ,4\pi]$. 
 \\
 
 \textbf{TYPE 6:-}
 \begin{eqnarray*}
 	y_i = \frac{sin(x_i)}{x_i} + \xi_i, ~~\xi_i\sim N[0,0.4]
 \end{eqnarray*}
 ~~~~~~~~~~ and  $x_i$ is from $U[-4\pi ,4\pi]$. 
 \\

 \textbf{TYPE 7:-}
 \begin{eqnarray*}
 	~y_i = ~ \left| \frac{x_i-1}{4}\right|+ \left| sin(\pi(1+\frac{x_i-1}{4})) \right | + 1 +\xi_i, 
 \end{eqnarray*}
 ~~~~~~~$\xi_i\sim U[-0.4,0.4]$~~ and  $x_i$ is from $U[-4\pi ,4\pi]$. 
 \\
 
 \textbf{TYPE 8:-}
 \begin{eqnarray*}
 	~~~~~~~  ~y_i = ~ \left| \frac{x_i-1}{4}\right|+ \left| sin(\pi(1+\frac{x_i-1}{4})) \right | + 1 +\xi_i,
 \end{eqnarray*}
 ~~~~~~$\xi_i\sim U[-0.6,0.6]$~~ and  $x_i$ is from $U[-4\pi ,4\pi]$. \\
 \\
 All datasets contain 100 training samples with noise and 500 non-noise testing samples. To avoid the biased comparison, ten independent groups of noisy samples were generated randomly in MATLAB (http://in.mathworks.com/) for all type of datasets.

 	\begin{table*}
 		\centering
 		\caption{Numerical results on Artificial Datasets}
 		\label{my-label}
 			\begin{tabular}{|l|l|l|l|l|l|l|l|l|l|}
 				\hline
 				&               & $\tau_2,~\tau_1$       & SSE/SST       & SSR/SST       & RMSE          & MAE           & Sparsity$\%$ & ( q,c, $\epsilon$  )   & \tiny{CPU time} \\\hline
 				
 				\multirow{4}{*}{\textbf{TYPE 1}}        & \multicolumn{2}{|c|}{$\epsilon$-SVR}  & 0.0166~$\pm$~0.0077 & 0.9145~$\pm$~0.0676 & 0.0408~$\pm$~0.0099 & 0.0328~$\pm$~0.0079 & 42.7     & ( 4,0.5,0.1 )   & 0.73     \\	\cline{2-10}
 				& RP-$\epsilon$-SVR & 2,~0.5    & 0.0155 $\pm$ 0.0053 & 0.9540~$\pm$~0.0509 & 0.0400~$\pm$~0.0080 & 0.0324~$\pm$~0.0069 & 32.4     & ( 4,0.5,0.1 )   & 2.35     \\
 				&               & 2,~0.6    & 0.0155~$\pm$~0.0056 & 0.9474~$\pm$~0.0517 & 0.0399~$\pm$~0.0085 & 0.0324~$\pm$~0.0073 & 27.8     & ( 4,0.5,0.1 )   & 2.26     \\
 				&               & 2,~0.4    & 0.0155~$\pm$~0.0059 & 0.9579~$\pm$~0.0526 & 0.0399~$\pm$~0.0086 & 0.0324~$\pm$~0.0074 & 44.7     & ( 4,0.5,0.1 )   & 2.50     \\\hline
 				\multirow{4}{*}{\textbf{TYPE 2}}     & \multicolumn{2}{|c|}{$\epsilon$-SVR}              & 0.0271~$\pm$~0.0152 & 0.9285~$\pm$~0.0849 & 0.0519~$\pm$~0.0125 & 0.0416~$\pm$~0.0085 & 65.7     & ( 4,1,0.2 )     & 0.83     \\ \cline{2-10}
 				& RP-$\epsilon$-SVR & 1,~0.1    & 0.0260~$\pm$~0.0148 & 0.9162~$\pm$~0.0839 & 0.0509~$\pm$~0.0124 & 0.0412~$\pm$~0.0075 & 59.9     & ( 4,1,0.2 )     & 2.44     \\ 
 				&               & 1,~0.2    & 0.0264~$\pm$~0.0150 & 0.9096~$\pm$~0.0801 & 0.0513~$\pm$~0.0126 & 0.0418~$\pm$~0.0085 & 59.8     & ( 4,1,0.2 )     & 2.43     \\
 				&               & 1.2,~ 0.1 & 0.0259~$\pm$~0.0133 & 0.9477~$\pm$~0.0828 & 0.0508~$\pm$~0.0122 & 0.0402~$\pm$~0.0082 & 57.2     & ( 4,1,0.2 )     & 2.58     \\\hline
 				\multirow{4}{*}{\textbf{TYPE 3}}   &  \multicolumn{2}{|c|}{$\epsilon$-SVR}            & 0.0382~$\pm$~0.0288 & 0.9818~$\pm$~0.0998 & 0.0604~$\pm$~0.0193 & 0.0471~$\pm$~0.0103 & 75.2     & ( 4,2,0.3 )     & 0.79     \\ \cline{2-10}
 				& RP-$\epsilon$-SVR & 1,~0.2    & 0.0377~$\pm$~0.0294 & 0.9665~$\pm$~0.0979 & 0.0598~$\pm$~0.0202 & 0.0464~$\pm$~0.0107 & 72.4     & ( 4,2,0.3 )     & 2.40     \\
 				&               & 1,~0.3    & 0.0381~$\pm$~0.0274 & 0.9554~$\pm$~0.0971 & 0.0605~$\pm$~0.0189 & 0.0488~$\pm$~0.0123 & 74.3     & ( 4,2,0.3 )     & 2.60     \\
 				&               & 2,~1.2    & 0.0376~$\pm$~0.0289 & 0.9663~$\pm$~0.0975 & 0.0598~$\pm$~0.0195 & 0.0472~$\pm$~0.0105 & 74.4     & ( 4,2,0.3 )       & 2.46     \\\hline
 				\multirow{4}{*}{\textbf{TYPE 4}}         &  \multicolumn{2}{|c|}{$\epsilon$-SVR}           & 0.0183~$\pm$~0.0091 & 0.8856~$\pm$~0.0402 & 0.0428~$\pm$~0.0093 & 0.0353~$\pm$~0.0084 & 44.5     & ( 4,0.5,0.1 )   & 0.73     \\ \cline{2-10}
 				& RP-$\epsilon$-SVR & 1.5,~0.3  & 0.0173~$\pm$~0.0095 & 0.9121~$\pm$~0.0648 & 0.0412~$\pm$~0.0108 & 0.0342~$\pm$~0.0093 & 57.2     & ( 4,0.5,0.1 )   & 2.50     \\
 				&               & 1.5,~0.2  & 0.0174~$\pm$~0.0096 & 0.9195~$\pm$~0.0667 & 0.0412~$\pm$~0.0112 & 0.0341~$\pm$~0.0101 & 44.6     & ( 4,0.5,0.1 )   & 2.43     \\
 				&               & 1.5,~0.1  & 0.0177~$\pm$~0.0095 & 0.9253~$\pm$~0.0704 & 0.0417~$\pm$~0.0108 & 0.0348~$\pm$~0.0096 & 43.4     & ( 4,0.5,0.1 )   & 2.41     \\\hline

 				\multirow{4}{*}{\textbf{TYPE 5}}       &  \multicolumn{2}{|c|}{$\epsilon$-SVR}                  & 0.1160~$\pm$~0.0498 & 0.8639~$\pm$~0.1561 & 0.1083~$\pm$~0.0220 & 0.0891~$\pm$~0.0184 & 36.7     & ( 4,0.5,0.2 )   & 0.79     \\ \cline{2-10}
 				& RP-$\epsilon$-SVR & 1.2,~0.2  & 0.1143~$\pm$~0.0523 & 0.8537~$\pm$~0.1544 & 0.1072~$\pm$~0.0232 & 0.0881~$\pm$~0.0190 & 41.6     & ( 4,0.5,0.2 )   & 2.81     \\
 				&               & 1.1,~0.2  & 0.1122~$\pm$~0.0528 & 0.8400~$\pm$~0.1519 & 0.1061~$\pm$~0.0237 & 0.0870~$\pm$~0.0202 & 37.8     & ( 4,0.5,0.2 )   & 2.58     \\
 				&               & 1.0,~0.2  & 0.1129~$\pm$~0.0508 & 0.8312~$\pm$~0.1520 & 0.1067~$\pm$~0.0225 & 0.0877~$\pm$~0.0190 & 43.2     & ( 4,0.5,0.2 )   & 2.58     \\\hline
 				
 				\multirow{4}{*}{\textbf{TYPE 6}}  &  \multicolumn{2}{|c|}{$\epsilon$-SVR}         & 0.1861~$\pm$~0.0858 & 0.7392~$\pm$~0.2029 & 0.1368~$\pm$~0.0277 & 0.1117~$\pm$~0.0240 & 28.3     & ( 4,0.125,0.2 ) & 0.77     \\ \cline{2-10}
 				& RP-$\epsilon$-SVR & 1.4,~0.1  & 0.1747~$\pm$~0.0758 & 0.7999~$\pm$~0.2096 & 0.1326$\pm$~0.0256 & 0.1072~$\pm$~0.0230 & 35       & ( 4,0.125,0.2 ) & 3.19     \\
 				&               & 1.3,~0.1  & 0.1749~$\pm$~0.0769 & 0.7731~$\pm$~0.1968 & 0.1327~$\pm$~0.0258 & 0.1081~$\pm$~0.0226 & 35.7     & ( 4,0.125,0.2 ) & 3.22     \\
 				&               & 1.2,~0.1  & 0.1749~$\pm$~0.0745 & 0.7439~$\pm$~0.1939 & 0.1330~$\pm$~0.0250 & 0.1078~$\pm$~0.0219 & 32.9     & ( 4,0.125,0.2 ) & 3.15     \\\hline
 				\multirow{4}{*}{\textbf{TYPE 7}}       & \multicolumn{2}{|c|}{$\epsilon$-SVR}          & 0.0166~$\pm$~0.0039 & 1.0206~$\pm$~0.0524 & 0.1256~$\pm$~0.0164 & 0.1013~$\pm$~0.0148 & 44       & ( 2,32,0.2 )    & 0.85     \\ \cline{2-10}
 				& RP-$\epsilon$-SVR & 0.5,~0.3  & 0.0159~$\pm$~0.0047 & 0.9971~$\pm$~0.0416 & 0.1226~$\pm$~0.0196 & 0.0989~$\pm$~0.0168 & 44       & ( 2,32,0.2 )    & 2.97     \\
 				&               & 0.5,~0.2  & 0.0162~$\pm$~0.0049 & 1.0047~$\pm$~0.0461 & 0.1237~$\pm$~0.0201 & 0.1002~$\pm$~0.0173 & 42       & ( 2,32,0.2 )    & 2.85     \\
 				&               & 1.2,~0.1  & 0.0162~$\pm$~0.0049 & 1.0047~$\pm$~0.0461 & 0.1236~$\pm$~0.0201 & 0.1001~$\pm$~0.0173 & 42       & ( 2,32,0.2 )    & 2.90     \\\hline
 				
 				\multirow{4}{*}{\textbf{TYPE 8}}         &  \multicolumn{2}{|c|}{$\epsilon$-SVR}          & 0.0286~$\pm$~0.0062 & 1.0325~$\pm$~0.0776 & 0.1650~$\pm$~0.0200 & 0.1334~$\pm$~0.0197 & 42       & ( 2,32,0.3 )    & 0.88     \\ \cline{2-10}
 				& RP-$\epsilon$-SVR & 0.8,~0.3  & 0.0273~$\pm$~0.0089 & 1.0248~$\pm$~0.0705 & 0.1603~$\pm$~0.0279 & 0.1295~$\pm$~0.0263 & 41       & ( 2,32,0.3 )    & 2.77     \\
 				&               & 0.7,~0.3  & 0.0273~$\pm$~0.0095 & 1.0198~$\pm$~0.0685 & 0.1600~$\pm$~0.0295 & 0.1296~$\pm$~0.0274 & 41       & ( 2,32,0.3 )    & 2.84     \\
 				&               & 0.6,~0.3  & 0.0275~$\pm$~0.0092 & 1.0173~$\pm$~0.0647 & 0.1607~$\pm$~0.0287 & 0.1300~$\pm$~0.0255 & 43       & ( 2,32,0.3 )    & 3.03     \\\hline
 			\end{tabular}
 	\end{table*}
 
  \begin{figure}
 	\centering
 	\begin{tabular}{c}
 		\subfloat[]{\includegraphics[width = 3.2in,height=1.8in]{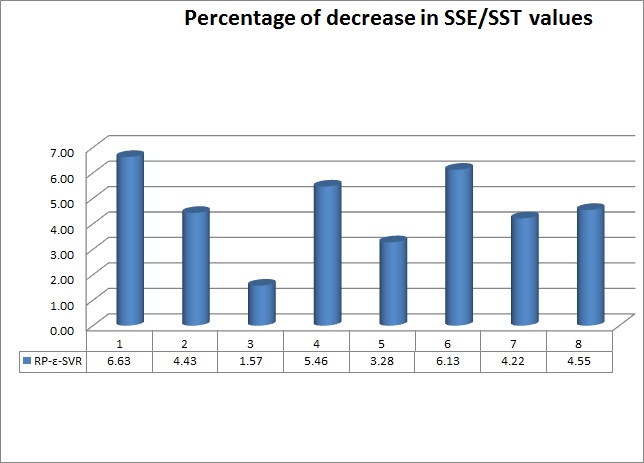}}\\
 		\subfloat[]{\includegraphics[width=3.2in, height=1.8in]{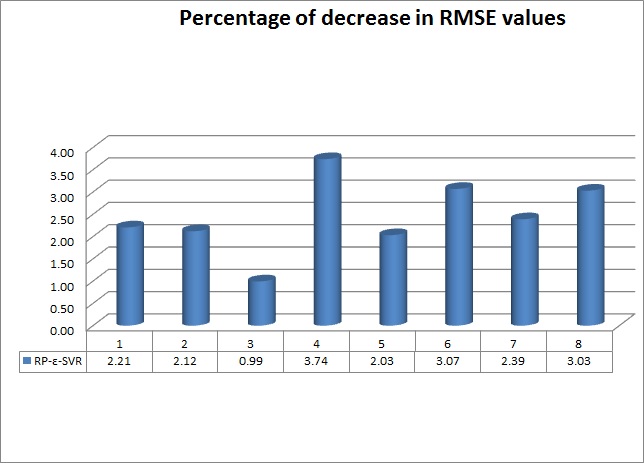}}
 	\end{tabular}
 	\begin{tabular}{c}
 		\subfloat[]{\includegraphics[width = 3.2in,height=1.8in]{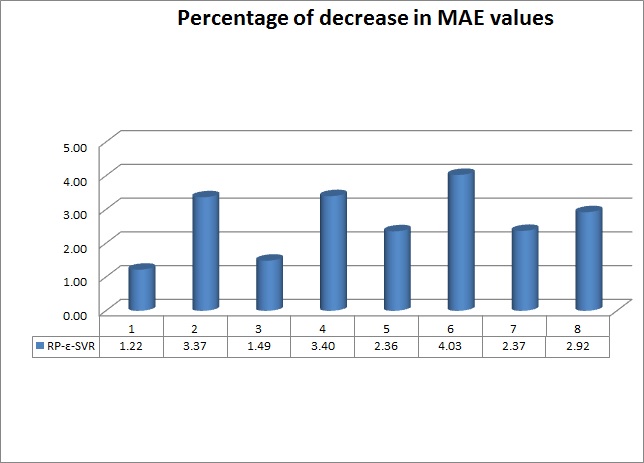}}\\
 		\subfloat[]{\includegraphics[width = 3.2in,height=2.0in]{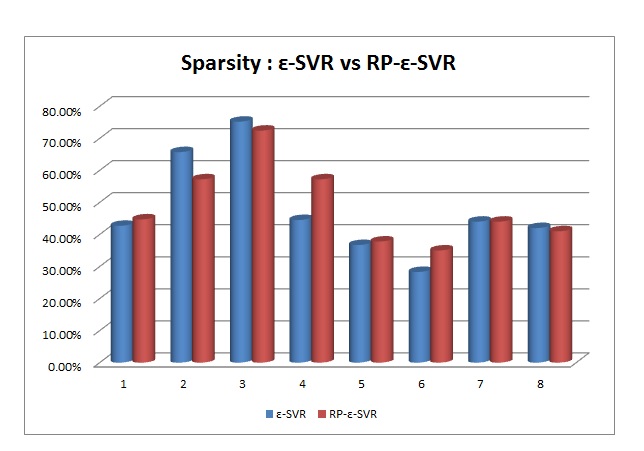}}
 	\end{tabular}
 	\caption{ Performance of the RP-$\epsilon$ SVR model over $\epsilon$-SVR model using different evaluation criteria on eight different artificial datasets listed in subsection (\ref{ad}) (represented by 1 to 8 on x-axis). }     
 	\label{comparision}                                         
 \end{figure}

 \begin{figure}
 	\centering
 	\begin{tabular}{c}
 		\subfloat[]{\includegraphics[width = 3.2in,height=1.9in]{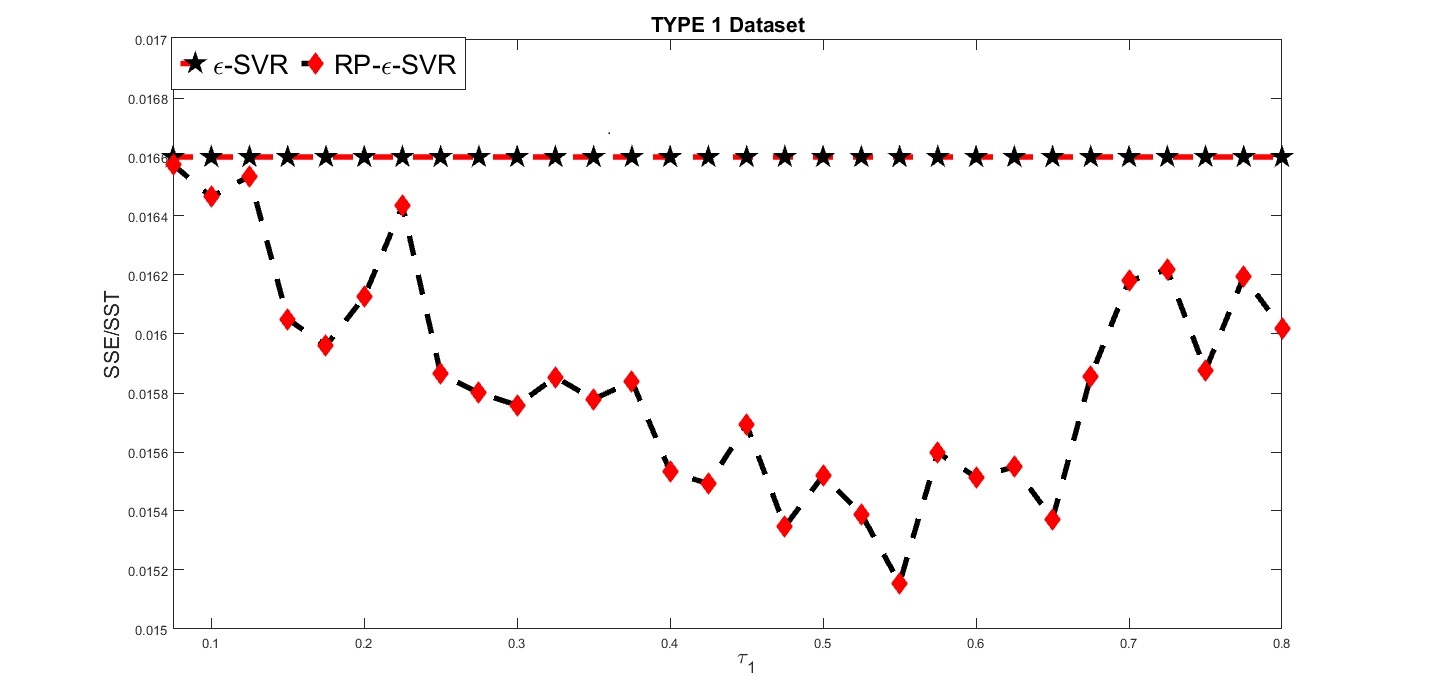}}\\
 		\subfloat[]{\includegraphics[width=3.2in, height=1.8in]{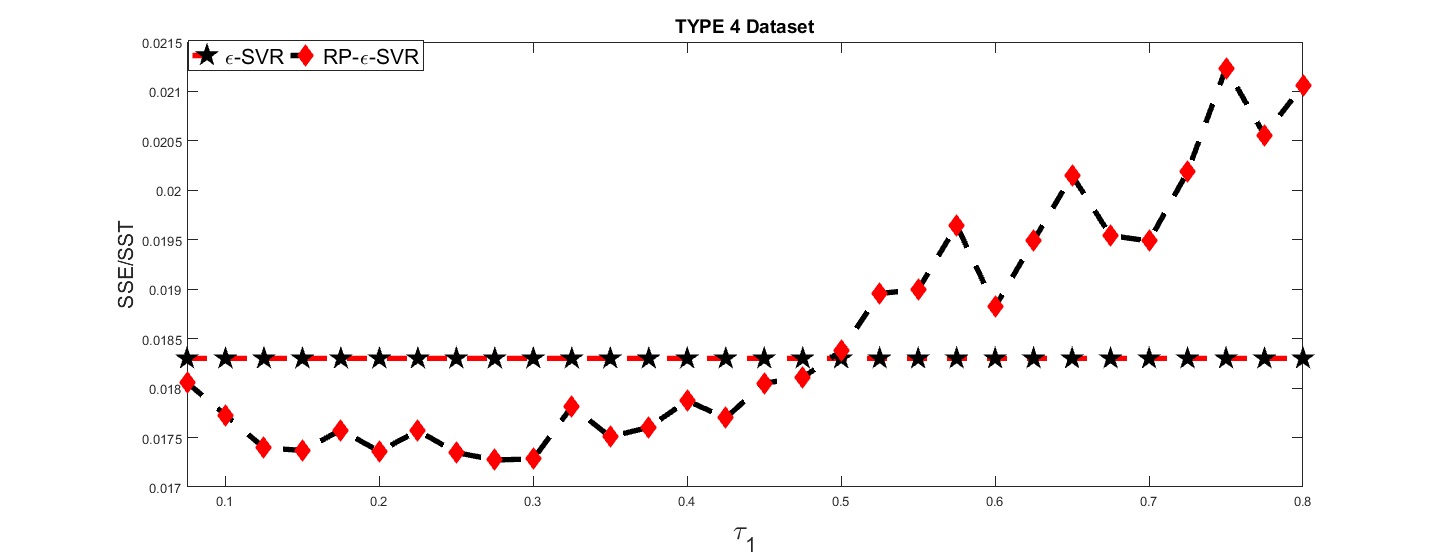}}
 	\end{tabular}
 	\begin{tabular}{c}
 		\subfloat[]{\includegraphics[width = 3.2in,height=2.0in]{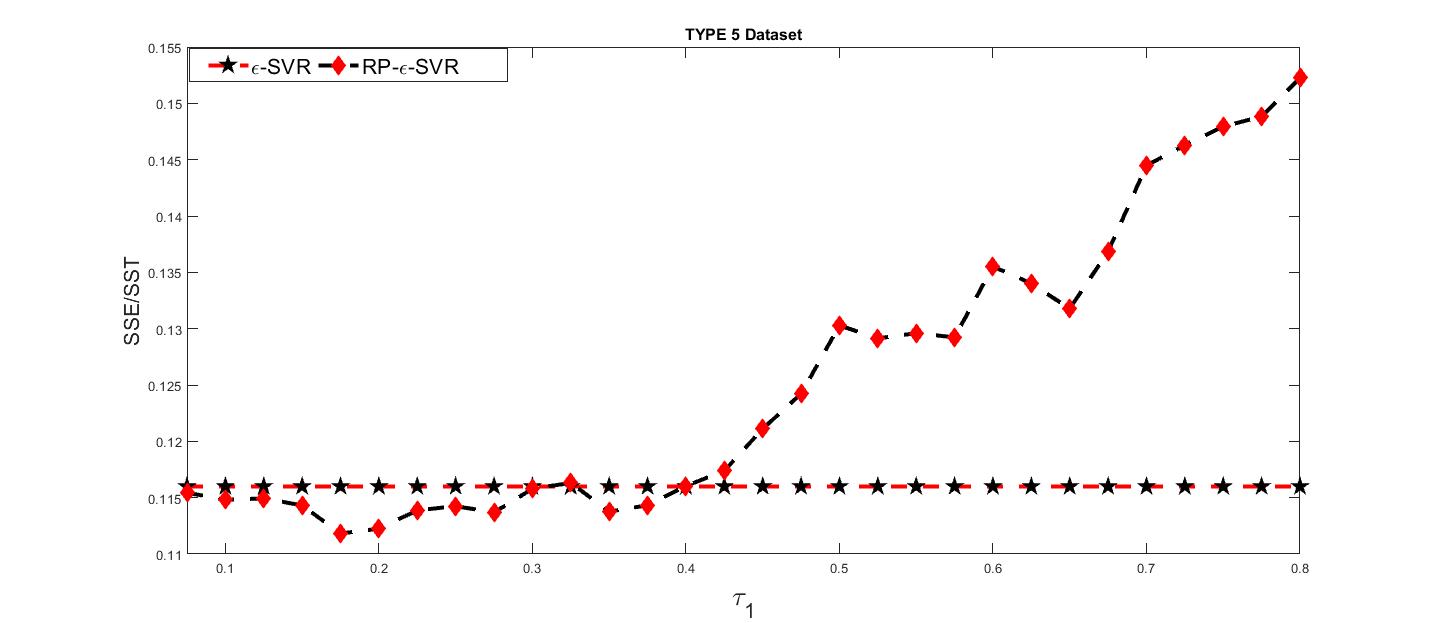}} \\
 		\subfloat[]{\includegraphics[width = 3.2in,height=2.0in]{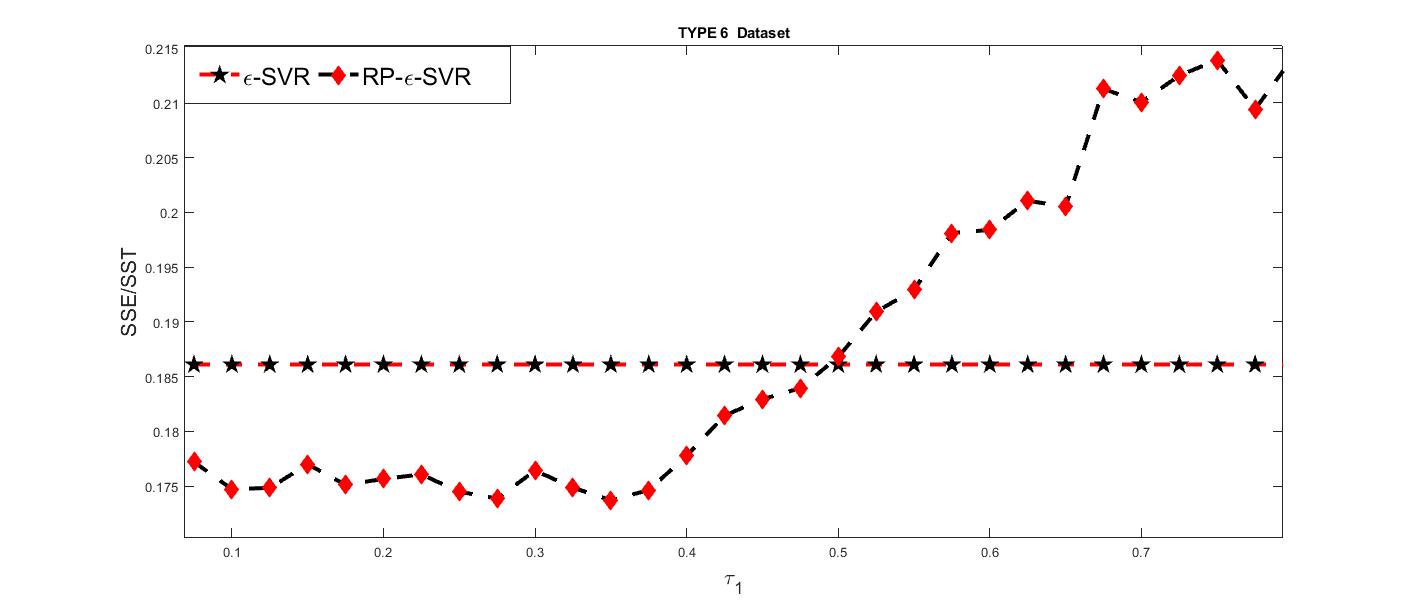}}
 	\end{tabular}
 	\caption{Plot of the SSE/SST values obtained by the RP-$\epsilon$-SVR model against the $\tau_1$ values on artificial datasets (a) TYPE 1  (b) TYPE 4 (c) TYPE 5 and (d) TYPE 6.}
 	\label{tau1}                                               
 \end{figure}

 Table \ref{my-label} lists the numerical results obtained from the experiments carried on the artificial datasets. We analyze the numerical results listed in Table \ref{my-label} as follows.
 
 \begin{enumerate}
 	\item [(a)]  The numerical results show that irrespective of the evaluation criteria and nature of noise present in the  artificial datasets, the proposed RP-$\epsilon$-SVR model always owns better generalization ability than existing $\epsilon$-SVR model. It also empirically verifies that proposed RP-$\epsilon$-SVR model is an improvement over the standard $\epsilon$-SVR model.  
 	\item [(b)] To realize this improvement, we have also computed the percentage of decrease in SSE/SST, RMSE and MAE values obtained by RP-$\epsilon$-SVR model over $\epsilon$-SVR model on artificial datasets as\\
 	percentage of the decrease in  value =\\~~~~~$~~~~~~~~~~~~~~~~~~~~~~~~~~\frac{\left(\mbox{RP-$\epsilon$-SVR value - }\epsilon-SVR ~ \mbox{value} \right)*100}{\epsilon-SVR ~\mbox{value}}$ . Figure \ref{comparision} shows the comparison and obtained improvement of the RP-$\epsilon$-SVR over the $\epsilon$-SVR using different evaluation criteria on eight artificial datasets. The use of  RP-$\epsilon$-SVR model over the $\epsilon$-SVR model always results significant improvement in the values of the SSE/SST, RMSE and MAE on artificial datasets. It is because of the fact that RP-$\epsilon$-SVR model can properly use the information of training set.  Figure \ref{comparision}(d)  compares the sparsity of the solution vector of the RP-$\epsilon$-SVR and $\epsilon$-SVR model.  We can realize that the sparsity of the solution vector of the RP-$\epsilon$-SVR is still comparable with the existing $\epsilon$-SVR model in the Table \ref{my-label}, though it can properly utilize the full information of the training set.
 	
 	\item [(c)] Table \ref{my-label} also lists the tunned parameters of the $\epsilon$-SVR model. The values of the parameter $C$,$q$ and $\epsilon$ of RP-$\epsilon$-SVR model have not been tunned explicitly. The tunned values of the $\epsilon$-SVR model has been only supplied to the RP-$\epsilon$-SVR model. It is noteworthy that, irrespective of the parameters values $C$, $\epsilon$ and $q$, tunned by the $\epsilon$-SVR model, the proposed RP-$\epsilon$-SVR model can find several $\tau_1$ and $\tau_2$ values on which it can outperform the $\epsilon$-SVR model. Figure \ref{tau1} shows the plot of the  SSE/SST values obtained using the proposed RP-$\epsilon$-SVR model against different $\tau_1$ values for a fixed value of the parameter $\tau_2$ on artificial datasets. It can be visualized that there exists several $\tau_1$ values for which the proposed  RP-$\epsilon$-SVR model obtains better SSE/SST values than $\epsilon$-SVR model. 
 \end{enumerate}
 
 \subsection{Benchmark datasets}                              
 For further evaluation, we have checked the performance of the proposed RP-$\epsilon$-SVR model on  UCI datasets namely, Yatch Hydro Dyanamics, Concrete Slump, Chwirut, Servo, Machine CPU, NO2, ENSO, Hahn1 and and AutoMpg. Yatch Hydro Dyanamics, Concrete Slump, Servo, Machine CPU, NO2, Autompg and Nelson datasets were downloaded from UCI repository \cite{UCIbenchmark} (archive.ics.uci.edu/ml). ENSO , Hahn1 and Nelson datsets were downoladed from $www.itl.nist.gov/div898/strd
 /nls/nls\_main.shtml $. For all the datasets, only feature vectors are normalized in the range of [0,1]. Ten-fold cross validation (Duda and Hart \cite{tenfold}) method has been used to report the numerical results for these datasets.
 
\begin{figure}
	\centering
	\begin{tabular}{c}
		\subfloat[]{\includegraphics[width = 3.0in,height=2.0in]{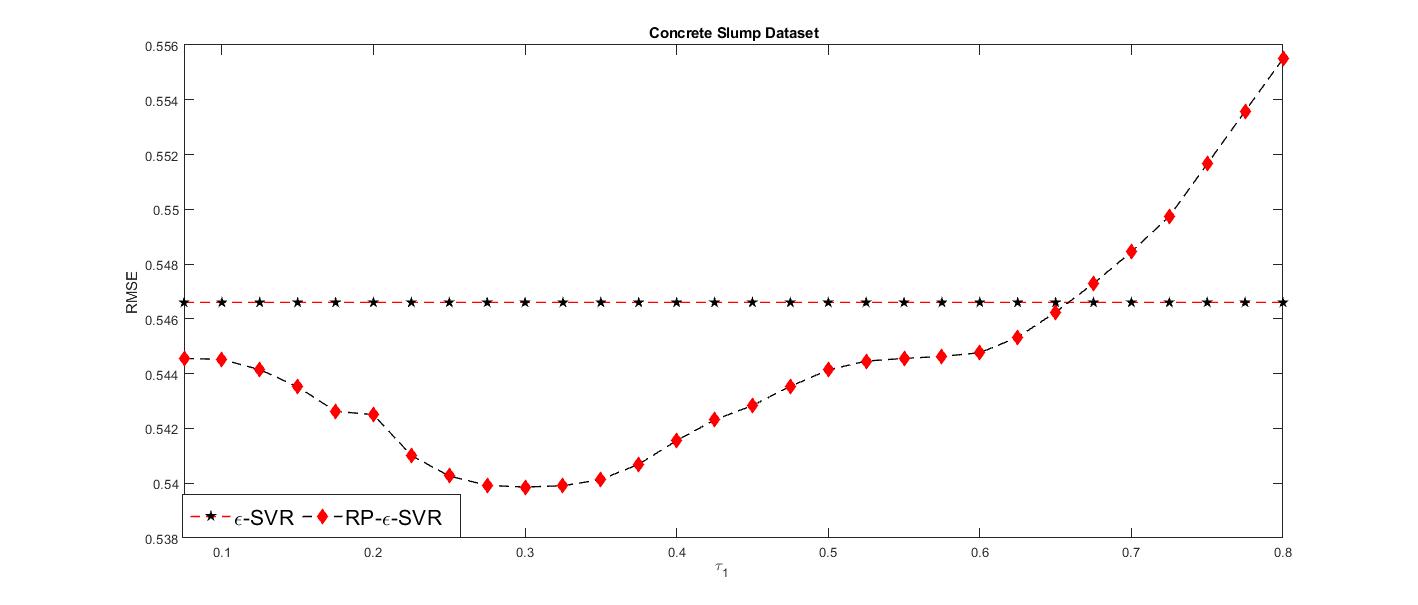}} \\
		\subfloat[]{\includegraphics[width=3.0in, height=2.0in]{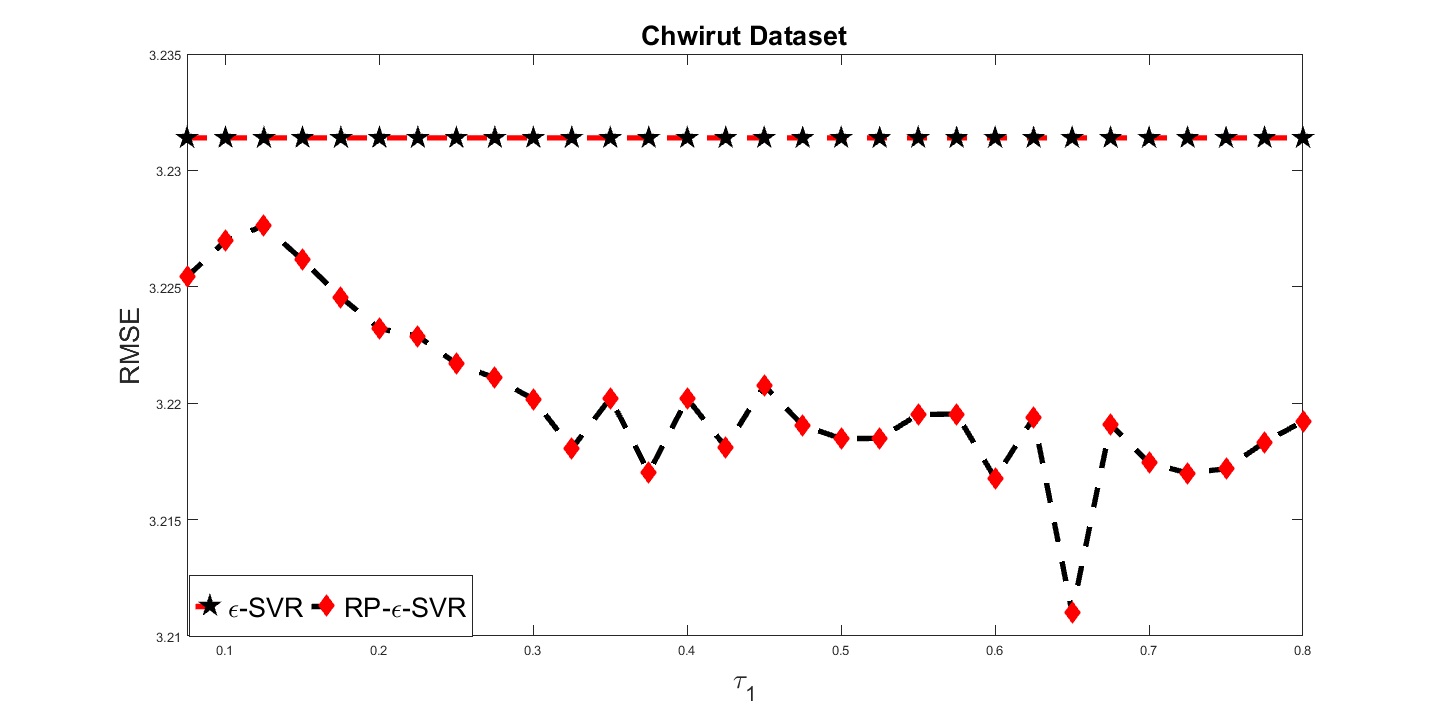}}
	\end{tabular}
	\begin{tabular}{c}
		\subfloat[]{\includegraphics[width = 3.0in,height=2.0in]{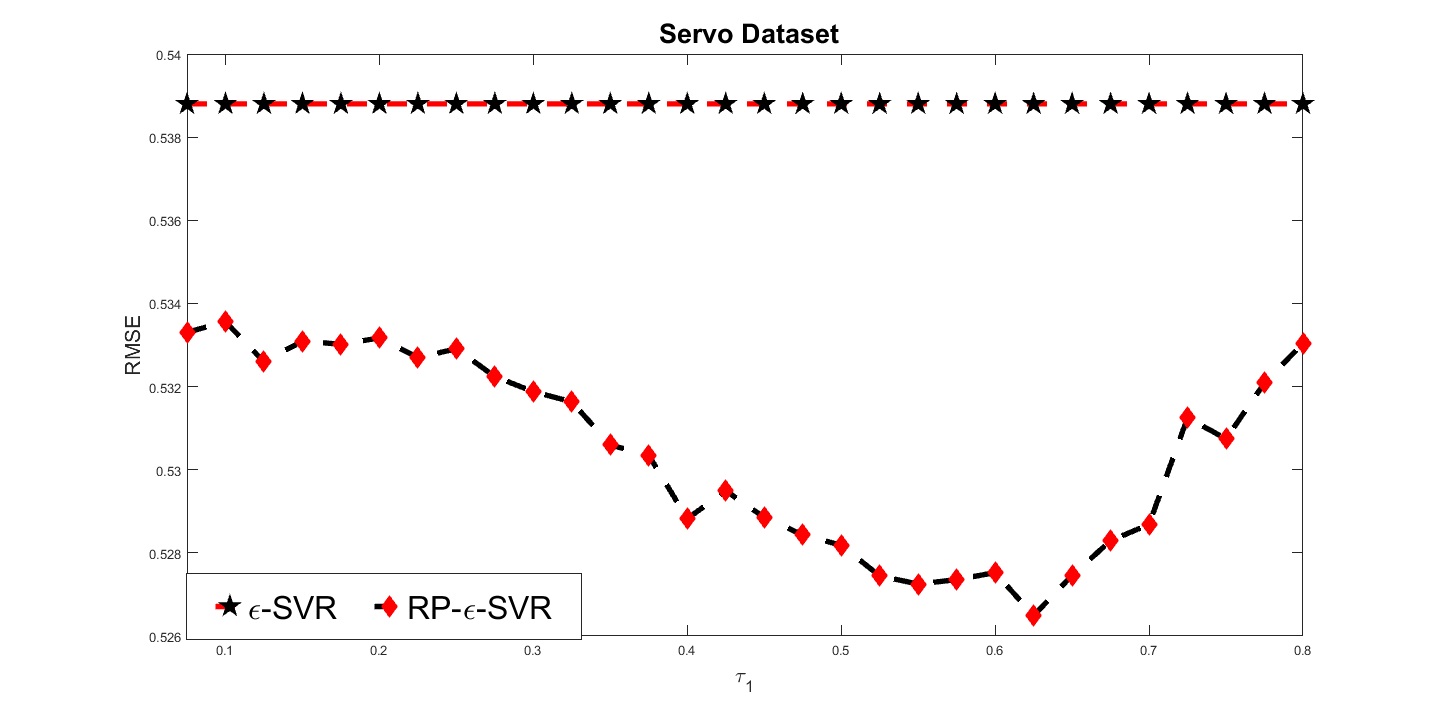}} \\
		\subfloat[]{\includegraphics[width = 3.0in,height=2.0in]{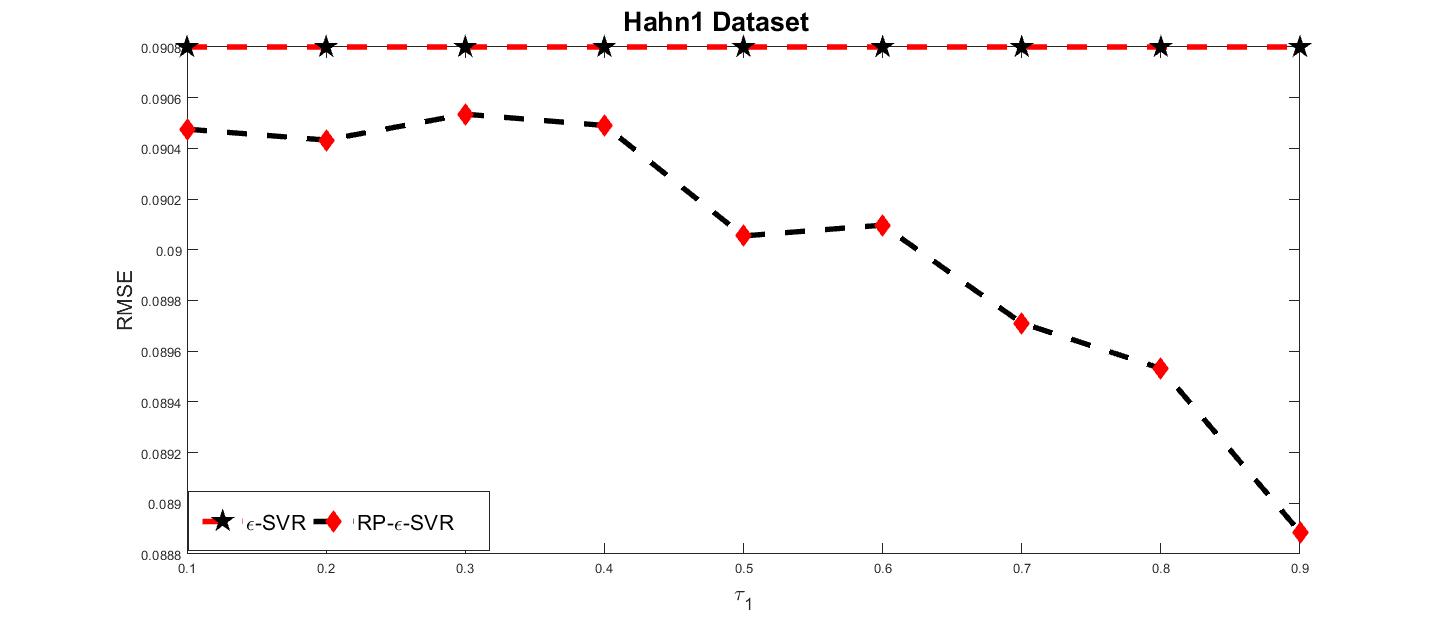}}
	\end{tabular}
	\caption{Plot of the RMSE values obtained by the RP-$\epsilon$-SVR model using different  $\tau_1$ values on (a) Concrete Slump (b) Chwirut  (c) Servo and (d) Hanh1 datasets.}          
	\label{UCIPLOT}                                     
\end{figure}

 Table \ref{UCI} lists the numerical results  obtained from the experiments carried on real-world benchmark datasets. The proposed RP-$\epsilon$-SVR always performs better than $\epsilon$-SVR model on several $\tau_1$ values on given datasets. The tunned parameters of the $\epsilon$-SVR method is also listed for different datasets. Similar to the line of the numerical results for artificial datasets, we can also analyze the numerical results listed in Table \ref{UCI} for benchmark datasets. Figure \ref{UCIPLOT} shows the plot of the RMSE values obtained by the proposed RP-$\epsilon$-SVR model against different $\tau_1$ values for the fixed value of the $\tau_2$ listed in the Table \ref{UCI} on UCI datasets. The proposed RP-$\epsilon$-SVR model can perform better than $\epsilon$-SVR model on several $\tau_1$ values as the RP-$\epsilon$-SVR model is more general model than $\epsilon$-SVR model. The best value of the $\tau_1$ is different  with datasets.     
 	\begin{table*}[h]
 		\centering
 		\caption{Numerical Results on UCI datasets}
 		\label{UCI}
 			{\fontsize{7.8}{7.8} \selectfont 
 				\begin{tabular}{|l|l|l|l|l|l|l|l|l|}
 					\hline
 					Dataset &           & $\tau_2,~\tau_1$ & SSE/SST          & SSR/SST          & RMSE                    & MAE              & ( q,c,$\epsilon$ )      & Sparsity$\%$ \\ \hline
 					\multirow{3}{*}{Yatch Hydro Dynamics} &  \multicolumn{2}{|c|}{$\epsilon$-SVR}    & 0.0034  $\pm$  0.0008 & 0.9809 $\pm$   0.0442 & 0.8633 $\pm$   0.2031        & 0.5166  $\pm$   0.1056 & ( 0.25,1024,0.2 ) & 34.34    \\ \cline{2-9}
 					&  & 2,~0.1            & 0.0027   $\pm$  0.0026 & 0.9868  $\pm$   0.0307 & 0.7194  $\pm$   0.3232        & 0.4185  $\pm$   0.1129 & ( 0.25,1024,0.2 ) & 41.66   \\
 					$~~~~~308 \times 7$ & RP-$\epsilon$-SVR               & 2,~0.2            & 0.0027   $\pm$  0.0026 & 0.9865   $\pm$  0.0305 & 0.7245   $\pm$  0.3198        & 0.4244   $\pm$  0.1135 & ( 0.25,1024,0.2 ) & 41.34     \\
 					&               & 2,~0.3            & 0.0028  $\pm$   0.0026 & 0.9866   $\pm$  0.0311 & 0.7371   $\pm$  0.3169        & 0.4335   $\pm$  0.1139 & ( 0.25,1024,0.2 ) & 40.69     \\\hline
 					
 					\multirow{3}{*}{Concrete Slump} &  \multicolumn{2}{|c|}{$\epsilon$-SVR} & 0.0073    $\pm$ 0.0049 & 0.9789  $\pm$   0.0711 & 0.5466   $\pm$  0.0608        & 0.4304  $\pm$   0.0664 & ( 2,1024,0.1 )    & 16.82  \\ \cline{2-9}
 					&  & 1.5,~0.3          & 0.0071  $\pm$   0.0053 & 0.9815   $\pm$  0.0576 & 0.5399  $\pm$   0.0763        & 0.4165  $\pm$   0.0668 & ( 2,1024,0.1 )    & 19.08      \\
 					$~~~~~103 \times 8$ & RP-$\epsilon$-SVR       &        1.5,~0.2          & 0.0072  $\pm$   0.0052 & 0.9820    $\pm$ 0.0577 & 0.5425  $\pm$   0.0768        & 0.4168  $\pm$   0.0694 & ( 2,1024,0.1 )    & 18.76    \\
 					&               & 1.5,~0.4          & 0.0072   $\pm$  0.0054 & 0.9809  $\pm$   0.0573 & 0.5416  $\pm$   0.0780        & 0.4183   $\pm$  0.0640 & ( 2,1024,0.1 )    & 18.44     \\\hline
 					
 					\multirow{3}{*}{Chwirut} &  \multicolumn{2}{|c|}{$\epsilon$-SVR}  & 0.0224 $\pm$   0.0118 & 0.9501 $\pm$    0.0640 & 3.2314 $\pm$    0.9571        & 2.2499   $\pm$ 0.5488 & ( 0.0156,64,0.3 ) & 11.00  \\ \cline{2-9}
 					
 					&  & 2,~0.7            & 0.0223 $\pm$   0.0122 & 0.9521$\pm$    0.0609 & 3.2174 $\pm$    0.9607        & 2.2421 $\pm$    0.5315 & ( 0.0156,64,0.3 ) & 11.05     \\
 					$~~214 \times 3$ & RP-$\epsilon$-SVR   & 2,~0.6            & 0.0223 $\pm$    0.0122 & 0.9509 $\pm$   0.0609 & 3.2168 $\pm$   0.9632        & 2.2373 $\pm$    0.5296 & ( 0.0156,64,0.3 ) & 10.95      \\
 					&                & 2,~0.5            & 0.0223   $\pm$ 0.0122 & 0.9506 $\pm$    0.0608 & 3.2185   $\pm$ 0.9638        & 2.2393 $\pm$    0.5306 & ( 0.0156,64,0.3 ) & 11.16     \\\hline
 					
 					\multirow{3}{*}{Servo} &  \multicolumn{2}{|c|}{$\epsilon$-SVR}   & 0.1651 $\pm$   0.1713 & 0.9125 $\pm$    0.4575 & 0.5388    $\pm$ 0.4261        & 0.3056  $\pm$   0.1536 & ( 0.125,4,0.1 )   & 36.59 \\ \cline{2-9}
 					
 					&  & 1,~0.2          & 0.1600 $\pm$   0.1650 & 0.8682 $\pm$   0.4208 & 0.5302 $\pm$   0.4168        & 0.3008  $\pm$  0.1476 & ( 0.125,4,0.1 )   & 37.59      \\
 					$~167 \times 5$ & RP-$\epsilon$-SVR   & 1,~0.3          & 0.1588  $\pm$  0.1622 & 0.8487  $\pm$  0.3995 & 0.5343 $\pm$   0.4199        & 0.3020  $\pm$  0.1483 & ( 0.125,4,0.1 )   & 37.79 \\
 					&               & 1,~0.4          & 0.1600  $\pm$  0.1631 & 0.8328  $\pm$  0.3906 & 0.5379 $\pm$    0.4244        & 0.3020 $\pm$    0.1502 & ( 0.125,4,0.1 )   & 37.39    \\\hline
 					
 					\multirow{3}{*}{Traizines} &  \multicolumn{2}{|c|}{$\epsilon$-SVR}   & 0.8497  $\pm$ 0.3690 & 0.4974 $\pm$   0.4741 & 0.1334  $\pm$   0.0274        & 0.0978    $\pm$ 0.0171 & ( 32 ,8,0.1 )     & 59.92    \\ \cline{2-9}
 					
 					&   & 1.9,~0.7          & 0.8243 $\pm$  0.2841  & 0.4834 $\pm$  0.3810  &  0.1328 $\pm$   0.0290       &  0.0972 $\pm$   0.0183 & ( 32 ,8,0.1 )     & 59.49    \\
 					$~186 \times 61$ & RP-$\epsilon$-SVR   & 1.8,~0.8            & 0.8410 $\pm$ 0.3129 & 0.4867 $\pm$   0.4212 &  0.1336 $\pm$  0.0283        & 0.0974 $\pm$   0.0174 & ( 32 ,8,0.1 )     &  59.20 \\
 					&               & 1.8,~0.7            & 0.8354 $\pm$   0.3046 & 0.4883 $\pm$    0.4094 & 0.1333 $\pm$    0.0286        & 0.0975 $\pm$    0.0178 & ( 32 ,8 ,0.1 )     & 59.86  \\\hline
 					
 					\multirow{3}{*}{Machine CPU } &  \multicolumn{2}{|c|}{$\epsilon$-SVR}     & 0.0192 $\pm$   0.0242 & 0.8887  $\pm$  0.1779 & 20.6659 $\pm$  21.6904       & 6.9929  $\pm$  5.1725 & ( 2,1024,0.1 )    & 6.75    \\ \cline{2-9}
 					&              & 1,0.05          & 0.0154 $\pm$   0.0162 & 0.9201 $\pm$    0.1615 & 18.2871 $\pm$   17.3716       & 6.5184 $\pm$   3.9406 & ( 2,1024,0.1 )    & 8.24  \\
 					$~209\times 8$ & RP-$\epsilon$-SVR & 1,~0.1            & 0.0164  $\pm$  0.0164 & 0.9114 $\pm$   0.1645 & 19.0037 $\pm$  17.6701       & 6.7139 $\pm$    3.9833 & ( 2,1024,0.1 )    & 7.76  \\
 					&    & 1,~0.2            & 0.0188  $\pm$  0.0176 & 0.8940 $\pm$    0.1716 & 20.5134  $\pm$ 18.6824       & 7.1097 $\pm$    4.1392 & ( 2,1024,0.1 )    & 6.53  \\ \hline

 					\multirow{3}{*}{NO2} &  \multicolumn{2}{|c|}{$\epsilon$-SVR}      & 0.4608  $\pm$  0.9955 & 0.6053  $\pm$  0.1810 & 0.4914  $\pm$ 0.0626     & 0.3882  $\pm$ 0.0469 & ( 0.5,8,0.3 )    & 41.71   \\ \cline{2-9}
 					
 					&  & 1.2,~0.1            & 0.4590   $\pm$  0.0794 & 0.5943  $\pm$   0.1736 & 0.4910  $\pm$  0.0586       & 0.3888  $\pm$   0.0452 & ( 0.5,8,0.3 )   & 47.98   \\
 					$~500\times 8$ & RP-$\epsilon$-SVR   & 1.2,~0.2            & 0.4602  $\pm$   0.0788 & 0.5919  $\pm$   0.1726 & 0.4919  $\pm$  0.0601       & 0.3891  $\pm$   0.0453  & ( 0.5,8,0.3 )    & 47.60   \\
 					&             & 1.5,~0.3            & 0.4595 $\pm$    0.0787 & 0.5975   $\pm$  0.1731 & 0.4914  $\pm$   0.0588       & 0.3886   $\pm$  0.0455  & ( 0.5,8,0.3 )    & 47.78  \\\hline
 					
 					\multirow{3}{*}{ENSO } &  \multicolumn{2}{|c|}{$\epsilon$-SVR}                  & 0.0072 $\pm$    0.0044 & 0.9861  $\pm$   0.0506 & 1.8174 $\pm$    0.4488        & 1.1529 $\pm$    0.2791 & ( 0.0625 ,128,0.2 )     & 37.49 \\ \cline{2-9}
 					
 					&  & 2,~0.2            & 0.0071 $\pm$    0.0044 & 0.9882 $\pm$   0.0509 &  1.8014  $\pm$  0.4426      & 1.1335  $\pm$  0.2720 & ( 0.0625 ,128, 0.2 )     & 38.16  \\
 					$~168\times 2$ & RP-$\epsilon$-SVR    & 2,~0.1           & 0.0071  $\pm$   0.0044 & 0.9880 $\pm$    0.0507 & 1.7998 $\pm$    0.4429        & 1.1321  $\pm$   0.2717 & ( 0.0625 ,128, 0.2 )     & 37.23    \\
 					&            & 1.9,~0.1            & 0.0071  $\pm$   0.0044 & 0.9881   $\pm$  0.0509 & 1.8008  $\pm$   0.4422        & 1.1332   $\pm$  0.2723 & ( 0.0625 ,128 , 0.2 )     & 38.02 \\\hline
 					
 					\multirow{3}{*}{Hahn1 } &  \multicolumn{2}{|c|}{$\epsilon$-SVR}                  & 0.0005 $\pm$    0.0010 & 1.0017  $\pm$   0.0142 & 0.0908 $\pm$    0.0134        & 0.0713 $\pm$    0.0137 & ( 0.0039 ,512,0.1 )     & 74.81 \\ \cline{2-9}
 					
 					&  & 1,~0.9            & 0.0005 $\pm$    0.0009 & 1.0018  $\pm$   0.0151 & 0.0889 $\pm$    0.0120        & 0.0702 $\pm$    0.0112 & ( 0.0039 ,512,0.1 )     & 75.80 \\ 
 					$~236\times 2$ & RP-$\epsilon$-SVR    & 1,~0.8           & 0.0005 $\pm$    0.0009 & 1.0017  $\pm$   0.0154 & 0.0895 $\pm$    0.0120        & 0.0708 $\pm$    0.0123 & ( 0.0039 ,512,0.1 )     & 75.52     \\
 					&            & 1,~0.7            & 0.0005 $\pm$    0.0009 & 1.0014  $\pm$   0.0149 & 0.0897 $\pm$    0.0127        & 0.0708 $\pm$    0.0135 & ( 0.0039 ,512,0.1 )     & 76.00  \\\hline
 					
 					\multirow{3}{*}{AutoMpg } &  \multicolumn{2}{|c|}{$\epsilon$-SVR}                  & 0.1153 $\pm$    0.0460 & 0.8929  $\pm$   0.0850 & 2.5669 $\pm$    0.4542        & 1.8444 $\pm$    0.2476 & ( 0.5 ,64,1 )     & 38.05 \\ \cline{2-9}
 					
 					&  & 1,~0.5            & 0.1124 $\pm$    0.0401 & 0.8628  $\pm$   0.0843 & 2.5517  $\pm$   0.4606        & 1.8430  $\pm$  0.2715 & ( 0.5 ,64,1 )     & 37.80  \\
 					$~398\times 9$ & RP-$\epsilon$-SVR    & 1,~0.2            & 0.1126  $\pm$   0.0412 & 0.8692 $\pm$    0.0878 & 2.5510 $\pm$    0.4648        & 1.8372  $\pm$   0.2723 & ( 0.5 ,64,1 )     & 38.27    \\
 					&            & 1,~0.1            & 0.1126  $\pm$   0.0412 & 0.8752   $\pm$  0.0909 & 2.5497  $\pm$   0.4520        & 1.8358   $\pm$  0.2704 & ( 0.5 ,64,1 )     & 38.89  \\\hline
 					
 					\multirow{3}{*}{Nelson } &  \multicolumn{2}{|c|}{$\epsilon$-SVR}                  & 0.1214 $\pm$    0.0745 & 0.9638  $\pm$   0.2003 & 1.2837 $\pm$    0.2927        & 0.9650 $\pm$    0.2249 & ( 0.0156 ,1024,0.2 )     & 6.94 \\ \cline{2-9}
 					
 					&  & 1,~0.1            & 0.1169 $\pm$    0.0633 & 0.9268  $\pm$   1.8887 & 1.2753  $\pm$   0.2839        & 0.9563  $\pm$  0.2148 & ( 0.0156 ,1024,0.2 )     & 8.33  \\
 					$~128\times 3$ & RP-$\epsilon$-SVR    & 1,~0.2            & 0.1171  $\pm$   0.0635 & 0.9259 $\pm$    0.1888 & 1.2753 $\pm$    0.2839        & 0.9545  $\pm$   0.2174 & ( 0.0156 ,1024,0.2 )     & 7.90    \\
 					&            & 1,~0.3            & 0.1185  $\pm$   0.0649 & 0.9268   $\pm$  0.1902 & 1.2831  $\pm$   0.2861        & 0.9589   $\pm$  0.2191 & ( 0.0156 ,1024,0.2 )     & 8.07  \\\hline                                  				
 			\end{tabular}}

 	\end{table*}

 	\begin{table*}[htp]
 		\begin{tabular}{|l|l|l|l|l|l|l|}
 			\hline
 			Dataset &  & SSE/SST & SSR/SST &  RMSE & MAE & CPU time \\ \hline
 			& $\epsilon$-SVR &  0.2316$~\pm~$0.0087  &  0.7860$~\pm~$ 0.0314 & 3.4315$~\pm~$ 0.3069 &  2.4030$~\pm~$ 0.2079 & 1.04\\ \cline{2-7}
 			Boston Housing & LS SVR &           0.2299$~\pm~$ 0.0120 & 0.8178$~\pm~$ 0.0328 &  3.4359$~\pm~$ 0.3104 & 2.4002 $~\pm~$ 0.2073 & 0.20 \\ \cline{2-7}
 			(350+112) $\times$ 14 & Huber SVR &  0.2438$~\pm~$ 0.0170 &  0.7052 $~\pm~$ 0.0258 &  3.5550$~\pm~$ 0.3278 & 2.4756$~\pm~$ 0.1833&     0.51     \\ \cline{2-7}
 			&  RP-$\epsilon$-SVR  &     0.2265$~\pm~$  0.0110 &   0.7929 $~\pm~$ 0.0344 &  3.4110$~\pm~$ 0.3218 &  2.4205$~\pm~$ 0.2195 & 12.93\\  \hline
 			
 			& $\epsilon$-SVR &  0.2339$~\pm~$0.0117  &  0.8617$~\pm~$ 0.0741 & 23.2254$~\pm~$ 3.5488 &  17.5270$~\pm~$2.4495 & 0.39\\ \cline{2-7}
 			Motorcycle & LS SVR &           0.2526$~\pm~$ 0.0334 & 0.7825$~\pm~$ 0.0530 &  23.6435$~\pm~$ 3.2028 & 18.1748 $~\pm~$ 2.2186 & 0.11 \\ \cline{2-7}
 			(100+33) $\times$  2 & $L_1$-Norm SVR &  0.2388$~\pm~$ 0.0141 &  0.9518 $~\pm~$ 0.0799 & 23.5480$~\pm~$3.6809 & 17.7823$~\pm~$ 2.4629&     1.39     \\ \cline{2-7}
 			&  RP-$\epsilon$-SVR  &     0.2335$~\pm~$  0.0121 &   0.8443 $~\pm~$ 0.0688 &  23.2805$~\pm~$ 3.6390 &  17.6877$~\pm~$ 2.6012 & 0.96\\  \hline                  
 			
 			& $\epsilon$-SVR & 0.6786$~\pm~$0.0298  &   0.5981$~\pm~$0.0233 &  0.6576$~\pm~$0.0305 &   0.4818$~\pm~$0.0171 & 9.70 \\ \cline{2-7}
 			Wine Quality (Red) 
 			&  LDMR &  0.6218$~\pm~$0.0031  & 0.4128$~\pm~$0.0081 &  0.6394$~\pm~$0.0148  & 0.5025$~\pm~$0.0101 &    10.23     \\ \cline{2-7}
 			(1000+599) $\times$ 14        &   Huber SVR  &     0.6689$~\pm~$0.0238 &  0.5745$~\pm~$0.0157  &   0.6550$~\pm~$0.0282  & 0.4913$~\pm~$0.0146 & 2.10\\ \cline{2-7}  
 			&   RP-$\epsilon$-SVR  &            0.6176$~\pm~$0.0030 & 0.5087$~\pm~$0.0112  &  0.6341$~\pm~$0.0163  & 0.4738$~\pm~$0.0119 & 261.80 \\  \hline                     
 		\end{tabular}
 		\caption{Comparision of performance of proposed RP-$\epsilon$-SVR model with different SVR models on UCI datasets}
 		\label{UCI2}
 	\end{table*}
 
 The proposed RP-$\epsilon$-SVR model is basically an improvement over popular and widely used $\epsilon$-SVR model. Therefore the numerical results presented in the Table \ref{UCI} compares the proposed RP-$\epsilon$-SVR model with the $\epsilon$-SVR model and are enough to empirically show that the proposed model is a better substitute of the $\epsilon$-SVR model. These numerical results also establishes the efficacy  of the proposed reward cum penalty loss function over existing  $\epsilon$-insensitive loss functions. 
 
 We  have also  compared  the performance of proposed RP-$\epsilon$-SVR model with some  other existing  traditional SVR models namely Huber SVR \cite{GUNNSVM} and LS-SVR\cite{LSSVR2}. Further, we have also  compared  the proposed RP-$\epsilon$-SVR model with some recent SVR models namely $L_1$-Norm SVR model\cite{l1normsvr} and LDMR model. The parameters of these models has  also been tuned using Exhaustive search method\cite{Exhaustivesearch} in their appropriate range.
 
 For the comparison, we have picked up three more UCI datasets namely Boston Housing, Motorcycle and Wine quality (Red). Datasets were partitioned into the  training set and testing set randomly ten times and  numerical results were reported by taking the mean and variance of the obtained  numbers. The cardinality of training set and testing set has been listed in the Table \ref{UCI2} . Table \ref{UCI2} also lists the comparison of the performance of the proposed  RP-$\epsilon$-SVR model  and other traditional and recent SVR models along with the CPU time.  It can be observed that the performance of the proposed RP-$\epsilon$-SVR model  is not only better than standard $\epsilon$-SVR model but, it also outperforms the other existing SVR models.

 \section{Conclusions}
 \label{sec6}
 This paper proposes a novel reward cum penalty loss function for handling the regression problem. Unlike the other existing loss functions, it can also take negative values. Like $\epsilon$-insensitive loss function, the reward cum penalty loss function not only penalizes data points which lie outside the $\epsilon$-tube of the regressor $f(x)$ but, it also assigns reward for the data points lying inside the $\epsilon$-tube. The trade-off between the reward and penalty can be controlled by the parameters $\tau_1$ and $\tau_2$.  The reward cum penalty loss function has been judiciously used in the proposed RP-$\epsilon$-SVR model in such a way that it can always obtain the sparse solution. The proposed RP-$\epsilon$-SVR model is a direct improvement over the standard $\epsilon$-SVR model as it can properly use the  full information of training set while preserving the robustness and sparsity of the solution. The standard $\epsilon$-SVR model is a particular case of the proposed RP-$\epsilon$-SVR model with choice of the parameters $\tau_2$ = 1 and $\tau_1$ = 0. Experimental results on several artificial and real world datasets show that the proposed RP-$\epsilon$-SVR model always owns better generalization ability than existing $\epsilon$-SVR model.

As compared to the standard $\epsilon$-SVR model, the RP-$\epsilon$-SVR model will be requiring to tune at least one extra parameter $\tau_1$. The parameter $\tau_2$  can be kept as constant and  parameter $C$ can be tunned appropriately instead. However, this extra tunning of parameter $\tau_1$ in RP-$\epsilon$-SVR model makes  its model selection time longer than $\epsilon$-SVR model.

 There are some potential problems for future studies.  We need  a development of the fast algorithm to solve the optimization problem of the proposed  RP-$\epsilon$-SVR model. It will make the RP-$\epsilon$-SVR model suitable for the large scale datasets.  A traversal algorithm for finding the best $\tau_1$ value in RP-$\epsilon$-SVR model is also required.

 \section*{Acknowledgement}
 This work was supported by the Ministry of Electronics and Information Technology Government of India under
 Visvesvaraya PhD Scheme for Electronics and IT Order No. Phd-MLA/4(42)/2015-16. We are also thankful to the Editor and the learned referee for their valuable comments.




\bibliography{final}
\end{document}